\documentclass[final,12pt]{clear2024} 

\title[Meaningful Causal Aggregation and Paradoxical Confounding]{Meaningful Causal Aggregation and Paradoxical Confounding}
\usepackage{times}
\usepackage{tikz}
\usepackage{mathtools}
\usepackage{wrapfig}
\usetikzlibrary{arrows.meta, arrows}
\usepackage{amsfonts}

\usepackage{amsfonts}
\usepackage{thmtools} 
\usepackage{thm-restate} 
\usepackage[capitalize,noabbrev]{cleveref}

\newtheorem{assumption}{Assumption}
\crefname{assumption}{assumption}{assumptions}
\Crefname{assumption}{Assumption}{Assumptions}

\newtheorem{thm}{Theorem}[section]
\crefname{thm}{theorem}{theorems}
\Crefname{thm}{Theorem}{Theorems}

\newtheorem*{rmk}{Remark}

\newcommand{\R}{{{\mathbb R}}}

\newcommand{\cA}{{{\mathcal A}}}

\newcommand{\cC}{{{\mathcal C}}}

\newcommand{\cE}{{{\mathcal E}}}

\newcommand{\cI}{{{\mathcal I}}}

\newcommand{\cN}{{{\mathcal N}}}

\newcommand{\cU}{{{\mathcal U}}}
\newcommand{\cV}{{{\mathcal V}}}

\newcommand{\cX}{{{\mathcal X}}}
\newcommand{\cY}{{{\mathcal Y}}}
\newcommand{\cZ}{{{\mathcal Z}}}

\makeatletter
\newcommand*\bigcdot{\mathpalette\bigcdot@{.5}}
\newcommand*\bigcdot@[2]{\mathbin{\vcenter{\hbox{\scalebox{#2}{$\m@th#1\bullet$}}}}}
\newcommand*\xbar[1]{%
  \;
  \hbox{%
    \vbox{%
      \hrule height 0.3pt 
      \kern0.3ex
      \hbox{%
        \kern-0.17em
        \ensuremath{#1}%
        \kern-0.05em
      }%
    }%
  }%
} 
\makeatother

\newcommand{\indep}{\perp \!\!\! \perp}

\newcommand{\defeq}{\vcentcolon=}

\newtheorem{innercustomthm}{Theorem}

\newtheorem{innercustomlemma}{Lemma}
\newenvironment{customlemma}[1]
  {\renewcommand\theinnercustomlemma{#1}\innercustomlemma}
  {\endinnercustomlemma}

\newtheorem{innercustomex}{Example}
\newenvironment{customex}[1]
  {\renewcommand\theinnercustomex{#1}\innercustomex}
  {\endinnercustomthm}

\clearauthor{%
 \Name{Yuchen Zhu} \Email{yuchen.zhu.18@ucl.ac.uk}\\
 \addr Department of Computer Science, University College London, UK
 \AND
 \Name{Kailash Budhathoki} \Email{kaibud@amazon.de}
 \AND
 \Name{Jonas K\"{u}bler} \Email{kuebj@amazon.de}
 \AND
 \Name{Dominik Janzing} \Email{janzind@amazon.de}\\
 \addr Amazon Research Tübingen, Germany%
}

\begin{document}

\maketitle

\begin{abstract}%
In aggregated variables the impact of interventions is typically ill-defined because different micro-realizations of the same macro-intervention can result in different changes of downstream macro-variables. We show that this ill-definedness of causality on aggregated variables can turn unconfounded causal relations into confounded ones and vice versa, depending on the respective micro-realization. We argue that it is practically infeasible to only use aggregated causal systems when we are free from this ill-definedness. 
Instead, we need to accept that macro causal relations are typically defined only with reference to the micro states.
On the positive side, we show that cause-effect relations can be aggregated when the macro interventions are such that the distribution of micro states is the same as in the observational distribution; we term this natural macro interventions. We also discuss generalizations of this observation.
\end{abstract}

\begin{keywords}%
  Causality, Causal Abstraction%
\end{keywords}
\vspace{-4pt}
\section{Introduction}
\vspace{-4pt}


High-stake topics in economics, politics, and health frequently hinge on aggregated variables like {\it employment rate}, {\it voter participation}, and {\it vaccination rate}. The aggregations often encompass entire business sectors or populations of a region. This prevalence motivates comprehending the aggregation of causal relations between such variables. Drawing from \citet{Chalupka2016MultiLevelCS,Rubenstein2017}, we term these aggregates as \textit{macro-variables} and the fine-grained variables as \textit{micro-variables}. We first identify three challenges of aggregating causal relations. 
\paragraph{Challenge 1. Aggregated causal relations can be ambiguous.} As shown in prior studies \citep{Rubenstein2017,spirtes_scheines_2004, beckers19}, consider the medical case of total cholesterol's (TC) influence on heart disease (HD). Here, TC, comprised of low-density lipoprotein (LDL) and high-density lipoprotein (HDL), is a macro-variable. Conflicting findings arise on TC's effect on HD risks. The discrepancy stems from LDL and HDL having inverse impacts on HD, leading to varied interpretations of TC based on LDL and HDL compositions.

\citet{Rubenstein2017} introduce exact transformation between macro- and micro-models. Yet, their criteria are stringent, demanding that every \textit{allowed} microscopic realization of a macro-intervention equally affects the aggregated downstream variables, which is a strong assumption on the set of interventions allowed. \citet{beckers19} extend this to even more robust consistency definitions, where equal effects are even required of \textit{all} potential microscopic realizations of the same macro-intervention. 

In cases impossible for exact transformation, \citet{Beckers2019ApproximateCA} introduce a family of methods to measure the difference between macro- and micro-models. However, this requires choosing weights over interventions on micro-variables. Given the high-dimensionality of micro-variables and the expansive realm of potential interventions, reaching a consensus on the weights is challenging for practitioners. Moreover, their discussion cannot be easily extended to address the subsequent challenge.

\paragraph{Challenge 2. Bridging macro intervention to micro implementation.} Practitioners need to realize high-level interventions on the low level, like actualizing a GDP boost and seeing how that impacts civil conflict \citep{Miguel2004EconomicSA}. 
One might adapt the idea of pre-specifying a distribution over micro-interventions from \citet[Definition 4.7]{Beckers2019ApproximateCA}, where a macro-intervention would be associated with a distribution over a family of micro-interventions. 
However, two issues arise: firstly, gaining consensus on the distribution, as highlighted in Challenge 1, is tough. Secondly, this yields a spread of micro-interventions that may not individually align with the intended macro-outcome, even if they do on average. The authors offer no clear guidance on how to actualize micro-interventions while maintaining consistency with the macro-level.

\paragraph{Challenge 3. The micro-model may not be known.} In some scenarios, only isolated facts about the micro-model are known. For example, one may know that the relation between product sales and the revenue generated are not confounded, but the exact micro-level causal model is not known. In this case, how should we implement a high-level intervention on the low level such that the high-level causal effects implied by the high-level causal model are realized?

Driven by the three challenges, we delve into causal claims of interventions on macro-variables, {\it relative to their implementations on the micro level}. To the best of our knowledge, this is a novel perspective of aggregating micro causal models. \citet{janzing2022phen} also explores ill-definedness of causality arising from ill-definedness of interventions, but not in the specific sense of aggregating micro-models.  We make the following contributions: 1) Beyond the existing discussion on ambiguity of macro-interventions, we uncover that even the concept of {\it confoundedness} is ill-defined without an appropriate treatment of causal aggregation. Given the central role of confoundedness in causal inference, we hope this will intensify existing interest in causal aggregation. 2) On small micro-level causal graphs, we propose the concepts of `confounding-inhibiting micro-realizations', which induces unconfoundedness in the macro-model with an appropriately defined micro-level implementation, and `natural micro-realization', which allows the practitioner to preserve unconfoundedness of the micro-level but without knowledge of the micro-causal model. 3) We extend the concept of `natural intervention' to larger graphs, and propose macro-backdoor adjustment.



\vspace{-8pt}
\section{Preliminaries: Macro- and micro-intervention}\label{sec:formulation}
\vspace{-5pt}
In this work, micro-level causal models are standard acyclic structural equation models, introduced by \citet{pearl09} and using notation from \citet{peters_elements}: 

\begin{definition}\label{def:micro}
A \textbf{micro causal model} is an acyclic structural causal model (SCM). A \textbf{structural causal model} (SCM) $M\defeq (\mathbf{S}, P_{\mathbf{N}})$ consists of a collection $\mathbf{S}$ of $d$ \textbf{(structural) assignments} 
\begin{align}
    X_j \defeq f_j(\text{\bf PA}_j, U_j), \;\; j=1,...,d,
\end{align}
where $\text{\bf PA}_j \subseteq \left\{X_1,\cdots, X_d\right\} \setminus \left\{X_j\right\}$ are called \textbf{parents of} $X_j$; and a joint distribution $P_{\mathbf{U}} = P_{U_1,\cdots, U_d}$ over the noise variables, which we require to be jointly 
independent. The corresponding causal DAG $G$ has nodes $X_j$ and arrows into $X_j$ from each of its parents. Note, that $X_j$ does {\it not} have to be one-dimensional.
\end{definition}
\vspace{-5pt}
Macro-variables arise from applying an aggregation map on the micro-variables:
\begin{definition}[aggregation map, macro-variables]\label{def:coarsening}
Given a random variable $X$ with range $\cX$, an \textit{aggregation map} is a surjective but non-injective function $\pi: \cX \rightarrow \bar{\cX}$. Define $\bar{X} = \pi(X)$ to be the \textit{macro-variable} of $X$ under $\pi$. 
\end{definition}
 Graphically, we represent aggregation by the graph
 $X \leftrightarrow \bar{X}$, where the bi-directed arrow visualizes
 the fact that $\bar{X}$ can be seen as effect of $X$ in the observational distribution, but the direction is reversed once we talk about interventions on $\bar{X}$. Intuitively, readers can think about total sales as a consequence of individual sales before intervention; during an intervention on total sales, its change is reflected in individual-level sales (so the arrow is reversed)\footnote{To understand this reversal, consider a company where the management sets guidelines of increasing total sales and the staff implement this on the micro level.}.


We will later consider separate aggregations at each potentially high-dimensional
node of the micro-causal model. Each $X_j$ will often be a vector in $\R^n$ and $\bar{X}_j$ the sum or average of all its components. In our examples we will mostly work with the sum aggregation map, so we
use the same symbol $\pi$ instead of $\pi_j$ 
for each node.
Given a distribution $P(X_1,\dots,X_n)$ over the micro-variables, defining $\pi_j$ for each node induces a joint distribution
$P(X_1,\dots,X_n,\pi(X_1,),\dots,\pi(X_n))$.



\begin{definition}[amalgamated graph]
Given a micro causal model and a set of aggregation maps, we define the amalgamated graph, $G^a\defeq (\cV, \bar{\cV}, \cE, \cC)$, where $V \in \cV$ if $V$ is a micro-variable, $\bar{V} \in \bar{\cV}$ if $\bar{V}$ is a macro-variable. Further, $U\rightarrow V \in \cE$ if $U$ and $V$ are micro-variables and $U$ is a parent of $V$, and $S \leftrightarrow \bar{S} \in \cC$ if $\bar{S}$ is the aggregation of $S$.
\end{definition}

The first consequence of aggregation is that the aggregated variables may no longer show a well-defined causal  relation:

\begin{example}\label{ex:violate_rw17}
Consider the cause-effect relation $X\to Y$, with $X=(X_1,X_2)$ and $Y=(Y_1,Y_2)$. 
For $j=1,2$, set $Y_j:=\alpha_j X_j$. Define
$\bar{X}:=\pi_X(X_1,X_2)=X_1+X_2$, and likewise 
$\bar{Y}:=\pi_Y(Y_1,Y_2)=Y_1+Y_2$.
Then, for $\alpha_1\neq \alpha_2$, the effect of setting $\bar{X}$ to $\bar{x}$ on $\bar{Y}$ is ill-defined without any further specification. 
This example violates the consistency condition set out by \citet{Rubenstein2017}.  Due to space limitation, we elaborate in Appendix~\ref{sec:rubenstein-example}.
\end{example}

In reality, $\alpha_1 \neq \alpha_2$ is generic.
Since the operation `setting $\bar{X}$ to $\bar{x}$' is not well-defined \textit{a priori}, we need to specify 
the distribution according to which we randomize the micro state (the
`joint manipulation' in
\cite{spirtes_scheines_2004}):


\vspace{-5pt}
\begin{definition}[macro intervention] 
A \textbf{macro-intervention} at $\bar{x}$, denoted $do(\bar{x})$, is simply a perfect intervention on the macro-variable $\bar{X}$.
Given an aggregation map, $\pi : \cX \rightarrow \bar{\cX}$, $do(\bar{x})$ is {\it associated} with a probability measure over $\cX$, denoted $P^{do}_{\pi, \bar{x}}(X)$, the \textbf{micro-realization}, such that $\pi(x) = \bar{x}\;\; \forall x \in \text{supp}(P^{do}_{\pi,\bar{x}}(X))$. When the aggregation map is clear in context, we write $P^{do}_{\bar{x}}(X)$. 
\end{definition}
\vspace{-10pt}
\begin{rmk}
    1) This simple construction represents a shift in perspective: by requiring a macro-intervention to associate with a \textit{choice of} micro-realization, the actual effect of a macro-intervention can be computed naturally, addressing Challenge 2. 2) This also allows us to ask the constructive question: what are the best micro-realizations with respect to a particular goal I have on the macro-level e.g. no confounding on the macro-level?
\end{rmk}

\vspace{-7pt}

\citet{Chalupka2016MultiLevelCS} also define a macro-level intervention via actual interventions on micro-variables, but similar to \citet{Rubenstein2017, beckers19, pmlr-v213-massidda23a}, they consider scenarios where the result is insensitive to the difference in microscopic realizations. 
As a further difference to their setting, we consider stochastic interventions~\citep{Correa_Bareinboim_2020} as well as deterministic, while they only consider the latter.

\begin{figure*}[t]
\begin{minipage}{.6\textwidth}
    \centering
	\begin{tikzpicture}[roundnode/.style={circle, draw=black!100, fill=black!15, very thick, minimum size=0.5mm}, normalnode/.style={circle, draw=black!0, very thick, minimum size=0.5mm}, unfillednode/.style={circle, draw=black!100, very thick, minimum size=0.5mm}, squarednode/.style={rectangle, draw=red!60, fill=red!5, very thick, minimum size=5mm}, arrowhead/.style={Triangle[length=3mm, width=2mm]}]
		\node[text width=6cm, text=blue] at (3.4, 1.9) {$M$};
	    \node[text width=3cm, font = {\small}] at (0.7, 0.9) {$\bigg\{X_i \sim P_i\bigg\}_{i=1,\cdots, N}$};
		\node[unfillednode] (x1) at (-1.2,-0.3) {$X_1$};
		\node[text width=1cm] at (-0.7, -1) {$\vdots$};
		\node[unfillednode] (xN) at (-1.2,-2) {$X_N$};
		\node[unfillednode] (xbar) at (-1.2, -4) {$\bar{X}$};
		
		\node[unfillednode] (y1) at (2.2,-0.3) {$Y_1$};
		\node[text width=1cm] at (2.6, -1) {$\vdots$};
		\node[unfillednode] (yN) at (2.2,-2) {$Y_N$};
		\node[unfillednode] (ybar) at (2.2, -4) {$\bar{Y}$};
		
		\draw[-{Triangle[length=3mm, width=2mm]}, line width=1pt] (x1) -- node[font = {\small}, above=3pt]{$f_1(x_1) = \alpha_1 x_1$} (y1);
		\draw[-{Triangle[length=3mm, width=2mm]}, line width=1pt] (xN) -- node[font = {\small}, above=3pt]{$f_N(x_N) = \alpha_N x_N$} (yN);
		
		\draw[blue, thick, rounded corners=5pt]
  (-2.3,-2.8) rectangle ++(5.7,4.4);
  		\draw[dashed, thick, rounded corners=5pt]
  (-2,-2.6) rectangle ++(1.3,2.9);
    	\draw[dashed, thick, rounded corners=5pt]
  (1.7,-2.6) rectangle ++(1.3,2.9);
        \draw[{Triangle[open, length=2mm, width=2mm]}-{Triangle[open, length=2mm, width=2mm]}, line width=1pt] (-1.2, -2.6) -- node[font = {\small}, left=2pt]{$\pi$} (xbar);
        \draw[{Triangle[open, length=2mm, width=2mm]}-{Triangle[open, length=2mm, width=2mm]}, line width=1pt] (2.2, -2.6) -- node[font = {\small}, right=2pt]{$\pi$} (ybar);

	\end{tikzpicture}
\end{minipage}
\begin{minipage}{.5\textwidth}
      \centering
	\begin{tikzpicture}[roundnode/.style={circle, draw=black!100, fill=black!15, very thick, minimum size=0.5mm}, normalnode/.style={circle, draw=black!0, very thick, minimum size=0.5mm}, unfillednode/.style={circle, draw=black!100, very thick, minimum size=0.5mm}, squarednode/.style={rectangle, draw=red!60, fill=red!5, very thick, minimum size=5mm}, arrowhead/.style={Triangle[length=3mm, width=2mm]}]
		\node[text width=6cm, text=blue] at (3.4, 1.9) {$M^{X\sim P^{do}_{\mathbf{X}|\bar{x}}}$};
	    \node[text width=3cm, text=red, font = {\small}] at (1, 0.9) {$\bigg\{\mathbf{X} \sim P^{do}_{\mathbf{X}|\bar{x}}\bigg\}$};
		\node[unfillednode] (x1) at (-1.2,-0.3) {$X_1$};
		\node[text width=1cm] at (-0.7, -1) {$\vdots$};
		\node[unfillednode] (xN) at (-1.2,-2) {$X_N$};
		\node[unfillednode] (xbar) at (-1.2, -4) {$\color{red}\bar{x}$};
		
		\node[unfillednode] (y1) at (2.2,-0.3) {$Y_1$};
		\node[text width=1cm] at (2.6, -1) {$\vdots$};
		\node[unfillednode] (yN) at (2.2,-2) {$Y_N$};
		\node[unfillednode] (ybar) at (2.2, -4) {$\bar{Y}$};

		\draw[-{Triangle[length=3mm, width=2mm]}, line width=1pt] (x1) -- node[font = {\small}, above=3pt]{$f_1(x_1) = \alpha_1 x_1$} (y1);
		\draw[-{Triangle[length=3mm, width=2mm]}, line width=1pt] (xN) -- node[font = {\small}, above=3pt]{$f_N(x_N) = \alpha_N x_N$} (yN);
		
		\draw[blue, thick, rounded corners=5pt]
  (-2.3,-2.8) rectangle ++(5.7,4.4);
  		\draw[dashed, thick, rounded corners=5pt]
  (-2,-2.6) rectangle ++(1.3,2.9);
    	\draw[dashed, thick, rounded corners=5pt]
  (1.7,-2.6) rectangle ++(1.3,2.9);
		\draw[-{Triangle[length=3mm, width=2mm]}, line width=1pt] (xbar) -- node[font = {\small}, left=2pt]{$\pi$} (-1.2, -2.6);
        \draw[{Triangle[open, length=2mm, width=2mm]}-{Triangle[open, length=2mm, width=2mm]}, line width=1pt] (2.2, -2.6) -- node[font = {\small}, right=2pt]{$\pi$} (ybar);\end{tikzpicture}
\end{minipage}
    \caption{\textbf{Left:} The micro-variable causal model $M$ is described by an SCM, which is shown within the blue frame. The aggregation map $\pi$ is applied to the micro-variables $X_1,\cdots, X_N$ and $Y_1, \cdots, Y_N$, both contained in the dashed frame. $\bar{X}$ and $\bar{Y}$ are the macro variables which arise from the aggregation map.
    \textbf{Right:} A macro-intervention $ P^{do}_{\pi, \bar{x}}(X)$ is shown. The intervention applied to the macro variable $\bar{X}:= \bar{x}$ is realised as a (perfect) stochastic intervention on $X$, note that all values of $\mathbf{x}$ supported by $P_{\mathbf{X}|\bar{X}:=\bar{x}}$ give rise to $\bar{x}$.
\label{fig:sales_rev_micro_macro}} 
\vspace*{-7mm}
\end{figure*}
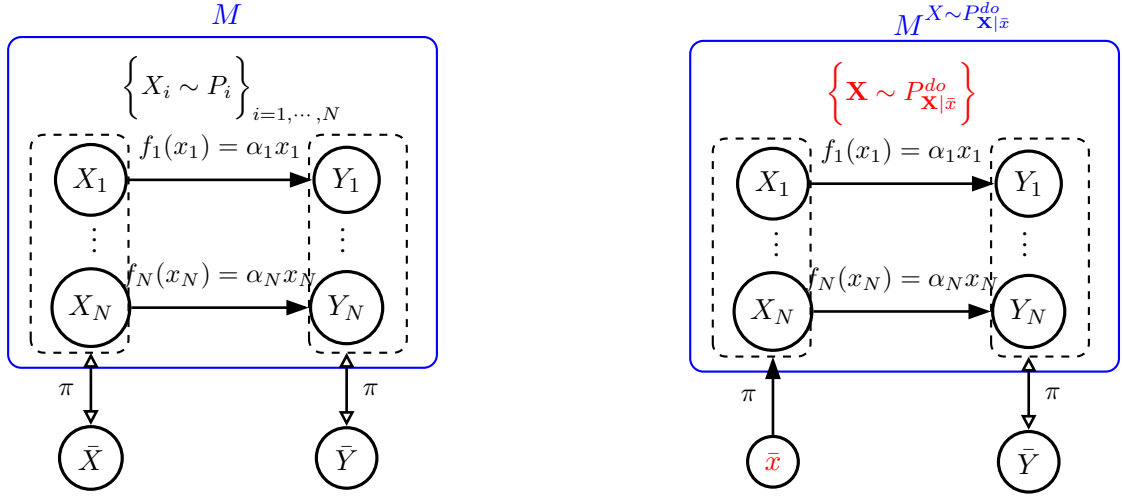


The following definition formalizes reversing the causal relation between $X$ and $\bar{X}$ with an intervention on $\bar{X}$, (see also Figure \ref{fig:sales_rev_micro_macro}):
\begin{definition}[macro-intervention graph]
Let $G^a = (\cV, \bar{\cV}, \cE, \cC)$ be an amalgamated graph. A macro intervention $P^{do}_{\pi, \bar{x}}$ maps $G^a$ to the \textit{macro-intervention graph}, ${G^{a}}' \defeq (\cV, \bar{\cV}, \cE', \cC')$, where $\cE' = \cE \bigcup \{\bar{X} \rightarrow X\} \setminus\{V \rightarrow X : V\in \cV\setminus\{X\}\}$ and  $\cC' = \cC \setminus \{X \leftrightarrow \bar{X}\}$.
\end{definition}

\vspace{-10pt}
\subsection{Confounding on the macro level} 
\vspace{-3pt}
Let $X$ and $Y$ be two (multi-variate) micro-variables with associated macro variables $\bar{X}$ and $\bar{Y}$. Suppose that $X$ precedes $Y$ in causal order.
Denote the observational distribution by $P$.
Given a macro intervention at $\bar{x}$, its associated micro-realization $ P^{do}_{\bar{x}}(X)$ ensures that the post-intervention quantity $P(\bar{y}|do(\bar{X}\defeq\bar{x}))$ is well-defined: $P(\bar{y}|do(\bar{X}\defeq\bar{x})) = \int_{\cX, \cY} P(\bar{y}|Y=y)P(y|do(X\defeq x))P^{do}_{\pi, \bar{x}}(x) dxdy$. We then define the notion of macro-confounding by comparing the observational and interventional distributions of macro variables, but first we restrict our discussion in this paper under one assumption:
\vspace{-3pt}
\begin{assumption}\label{assp:no_link_among_x}
    We focus on the case where there is no directed causal paths among variables to be aggregated.
\end{assumption}
\vspace{-3pt}
This is a simplifying assumption to make discussion on defining macro-confounding easier. This allows us to say that an intervention on the aggregated variable should not affect the noise on the macro-level, which arises under a change-of-coordinate of the micro-variable space. 

\begin{definition}[macro confounding]
Let $X$ and $Y$ be two micro-variables with associated macro variables $\bar{X}$ and $\bar{Y}$ and let $X$ precede $Y$ in causal order.
Let $\cI$ be a set containing macro-interventions, one for each $\bar{x} \in \text{Supp}[P(\bar{X})]$.
We say there is $\textit{macro-confounding}$ between macro variables $\bar{Y}$ and $\bar{X}$ if $P(\bar{Y}|\bar{X}=\bar{x}) \neq P(\bar{Y}|do(\bar{X}\defeq \bar{x}))$ for some $\bar{x}$. 
\end{definition}
\vspace{-10pt}
One may question whether it is reasonable to define confounding by disagreement of interventional and observational distributions? We can justify it from the angle of a coordinate change: aggregation can be thought of as a coordinate transformation of the space of micro-variables, where one of the new coordinates is given by the aggregation map. Recall Assumption~\ref{assp:no_link_among_x}, there are no directed causal paths among the variables being aggregated, hence there are also no directed causal paths among the variables under the new coordinates, then an intervention on $\bar{X}$ should keep the other variables fixed. It is therefore natural to think of confounding as the correlation between $\bar{X}$ and other variables under the new coordinate system in the observational distribution, but which can be broken in the interventional. This is captured by $P(\bar{Y}|\bar{X}=\bar{x}) \neq P(\bar{Y}|do(\bar{X} \defeq \bar{x}))$. We provide a detailed discussion of linear change-of-coordinate in Appendix~\ref{app:change-of-cood} and the nonlinear case in Section~\ref{sec:nonlinear-nongaussian}.

\vspace{-5pt}
\begin{definition}
A \textbf{confounding-inhibiting micro-realization} of a macro-intervention has the post-interventional distribution $P(\bar{Y}|do(\bar{X}\defeq\bar{x}))$ equal to the observational distribution $P(\bar{Y}|\bar{X}=\bar{x})$. On the other hand, a \textbf{confounding-inducing micro-realization} has $P(\bar{Y}|do(\bar{X}\defeq \bar{x})) \neq P(\bar{Y}|\bar{X}=\bar{x})$.
\end{definition}
\vspace{-15pt}

\section{Confounding is ambiguous} \label{subsec:aggre}
\vspace{-3pt}
Now we illustrate that confoundedness in the macro-level is ambiguous. We first discuss this in an unconfounded micro-model, where the macro-model may appear confounded for the simple observation that there are distinct micro-realizations to the same macro intervention. However, less obvious is the case even where the micro-model is {\it confounded}, there still exist cases where the micro-realizations can be chosen so that the macro-model appear unconfounded. 

All proofs are found in Appendix~\ref{app:proofs}. 

\vspace{-5pt}
\begin{rmk}
    We focus mostly on linear Gaussian models, but provide preliminary results to nonlinear, non-Gaussian, and discrete cases. We leave for future work a full treatment of the general case. 
\end{rmk}
\vspace{-10pt}
\subsection{Macro-confounding in unconfounded micro-models}\label{sec:unconfounded}

For simplicity of illustration, we consider a two-variable example, but the results can be readily extended to an $n-$variable linear setting.

To match the scenario in Example \ref{ex:violate_rw17}, 
suppose there are two identical shops, $A$ and $B$, each with two products, indexed $1$ and $2$, with prices $\alpha_1$ and $\alpha_2$, respectively. The sales $X_1,X_2$ of the two products are assumed to be independent Gaussians with means
$\mu_1,\mu_2$
 and variances $\sigma_1^2,\sigma_2^2$.
The total revenue, $\bar{Y}$ is given by $\bar{Y} = \alpha_1 X_1 + \alpha_2 X_2$. Now suppose the shop owners want to understand how total revenue changes \textit{w.r.t.} total sales, $\bar{X} = X_1 + X_2$. 

Shop $A$ owner performs the micro-realization $P^{do}(X_1,X_2|\bar{x}) \defeq P(X_1,X_2|\bar{x})$. Clearly this leads them to conclude that $\bar{X}, \bar{Y}$ is unconfounded, since 
\[P(\bar{Y} | do(\bar{X} \defeq \bar{x})) = P(\bar{Y} | x_1, x_2) P^{do}(x_1, x_2 | \bar{x}) = P(\bar{Y} | x_1, x_2) P(x_1, x_2 | \bar{x}) = P(\bar{Y} | \bar{x}). \] 
Consequentially, the shop  $A$ owner concludes on the following structural equation which is what would be obtained by regressing $\bar{Y}$ on $\bar{X}$:
\begin{align}
    \bar{Y} &= \frac{{\rm Cov}(\bar{Y}, \bar{X})}{{\rm Cov}(\bar{X}, \bar{X})}\bar{X} + N \label{eq:200}
    = \frac{\alpha_1 \sigma^2_1 + \alpha_2\sigma^2_2}{\sigma_1^2 + \sigma_2^2 } \bar{X} + N,
\end{align}
with     $N \indep \bar{X}$.  Meanwhile, shop $B$ owner performs an experiment also setting the total sales to $\bar{x}$, but they do this by observing how many items were sold for product $1$, and then turn up (or down) the advertising to make sure they sell $\bar{x} - x_1$ items for product $2$. This amounts to the macro intervention:
\vspace{-1ex}
\begin{align}
    P^{do}_{\bar{x}} (X_1, X_2) &= \cN \left(\begin{pmatrix}\mu_1 \\ \bar{x} - \mu_1\end{pmatrix}, \begin{pmatrix}\sigma_1^2 & -\sigma_1^2 \\ -\sigma_1^2 & \sigma_1^2\end{pmatrix}\right) \label{eq:toy_ex_2}
\end{align}
Shop $B$ owner, making this intervention, would instead conclude that the structural equation generating $\bar{Y}$ is 
\vspace{-1ex}
\begin{align}
    \bar{Y} &= \alpha_2 \bar{X} + \tilde{N}, \label{eq:201}
\end{align}
where $\tilde{N} = (\alpha_1 - \alpha_2)X_1$. Thus, whenever $\alpha_1 \neq \alpha_2$, we have $\tilde{N} \not \indep \bar{X}$ under $P$. 

Note that equations \eqref{eq:200} and \eqref{eq:201} generate the same observational distribution $P(\bar{Y}, \bar{X})$, but they correspond to different micro-realizations on $\bar{X}$, the former being \textit{confounding-inhibiting} while the latter being \textit{confounding-inducing}. This result can also be viewed from the perspective of coordinate change, which we discuss in Appendix~\ref{app:change-of-cood}.

\subsubsection{A simple positive result}\label{subsec:positive}  

\begin{wrapfigure}{r}{0.4\textwidth}
    \centering
\resizebox{0.3\textwidth}{!}{%
	\begin{tikzpicture}[roundnode/.style={circle, draw=black!100, very thick, minimum size=0.5mm}, greynode/.style={circle, draw=black!100, fill=black!15, very thick, minimum size=0.5mm}, normalnode/.style={circle, draw=black!0, very thick, minimum size=0.5mm}, squarednode/.style={rectangle, draw=red!60, fill=red!5, very thick, minimum size=5mm}, arrowhead/.style={Triangle[length=3mm, width=2mm]}]
		\node[roundnode] (x) at (-0.8,1) {$X$};
		\node[roundnode] (xbar) at (-0.8, -1) {$\bar{X}$};
		\node[roundnode] (y) at (1.8, 1) {$Y$};
		\node[roundnode] (ybar) at (1.8, -1) {$\bar{Y}$};
		\draw[{Triangle[open, length=2mm, width=2mm]}-{Triangle[open, length=2mm, width=2mm]}, line width=1pt] (x) -- node[font = {\small}, left=2pt]{$\pi_X$} (xbar);
		\draw[{Triangle[open, length=2mm, width=2mm]}-{Triangle[open, length=2mm, width=2mm]}, line width=1pt] (y) -- node[font = {\small}, right=2pt]{$\pi_Y$} (ybar);
		\draw[-{Triangle[length=3mm, width=2mm]}, line width=1pt] (x) -- (y);
  
  		\node[roundnode] (xbar) at (-0.8,-2.5) {$\bar{X}$};
		\node[roundnode] (ybar) at (1.8, -2.5) {$\bar{Y}$};
		\draw[-{Triangle[length=3mm, width=2mm]}, line width=1pt] (xbar) -- (ybar);
	\end{tikzpicture}
}
    \caption{With an appropriate microscopic micro-realizations of
    interventions on $\bar{X}$, the cause-effect relation (top) remains unconfounded on the macro level (lower).}
    \label{fig:ce}
    \vspace{-40pt}
\end{wrapfigure}
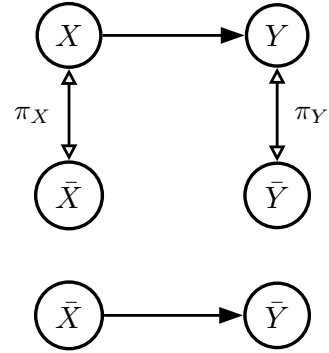

The following result provides a simple sufficient condition 
for which the unconfounded cause-effect relation in Figure \ref{fig:ce}, top,
turns into the unconfounded cause-effect relation in Figure \ref{fig:ce}, lower:

\begin{thm}\label{thm:main} 
Let $X\to Y$ and $\bar{X}$ and $\bar{Y}$ be macro-variables of $X$ and $Y$, respectively. Then the micro-realization
$P^{do}_{\bar{x}}(X) \defeq P(X|\bar{X} \defeq \bar{x})$ is confounding-inhibiting, i.e.
the graph $\bar{X} \to \bar{Y}$ 
can be drawn. 
\end{thm}

One should appreciate that 
this result shows an option for consistently aggregating cause-effect relations provided one is willing to accept its  {\it context dependence}: the relation  $\bar{X} \to \bar{Y}$  
holds whenever one assumes that the micro-realization $P^{do}_{\bar{x}}(X)$ is just the conditional distribution of micro states $X$ as is usual `in the wild'. This view, however, entails that we refer to different micro-realizations in different marginal distributions of $X$, because conditioning macro states would then in general give rise to different distributions of micro states. 

Nevertheless, adopting the distribution of micro states from the observational distribution seems like a natural definition for $P^{do}_{\bar{x}}(X)$
whenever there is no additional knowledge about the system telling us 
how a `more natural' distribution would look like.\footnote{Motivated by thermodynamics \citep{Balian1},
one may alternatively want to consider the distribution with maximal entropy subject to the constraint $\bar{X}=\bar{x}$ as `most natural'. However, entropy of continuous variables
implicitly refers to a reference measure (like the volume in phase space in physics), which again introduces an ambiguity. 
Nevertheless, {\it equilibrium} thermodynamics \citep{Adkins1983} provides
an example where `natural micro-realizations' exist: 
regardless of whether macroscopic variables like volume or pressure have been {\it observed} or {\it set} by an intervention, in both ceases the distribution of micro states is the
uniform distribution (maximum entropy) in the sub-manifold of the phase space satisfying the respective macro constraint.} 

\begin{definition}[Natural micro-realization]
    We call $P^{do}_{\bar{X}}(X) = P(X|\bar{X})$ the {\it natural micro-realization}\footnote{In practice, since this is stochastic, how to implement this can still be non-obvious, which is why we also introduce a deterministic analogue in Appendix~\ref{app:shift_intervention}, namely shift interventions. In Appendix~\ref{app:shift-change-of-coord}, we point out that one can realize the same structural coefficient as the natural intervention using shift interventions.}. 
\end{definition}

\begin{definition}[Natural macro-intervention]
    A natural macro-intervention of an aggregated variable is one that is associated with the natural micro-realization.
\end{definition}

\begin{rmk}
    Note, that the natural macro-intervention can be defined without reference to the micro-causal model. One only has to know that the underlying micro model does not confound $X$ and $Y$. This addresses Challenge 3 for cases with no micro-level confounding. 
\end{rmk}

\subsubsection{Non-Gaussian and non-linear generalization} \label{sec:nonlinear-nongaussian}
We can view $\bar{X}$ as the first coordinate after a {\it non-linear} coordinate transformation.
We will see that  even in {\it linear} models, noise terms that have a non-linear effect on the target may be more interpretable. Further, the most interpretable interventions may not necessarily come from {\it linear} coordinate changes. 

To this end, assume that 
$Y_j$ is given by
$
Y_j = f_j(X_j) + V_j, \quad j=1,\dots,N,
$
where $f_j$ may be non-linear functions and
$V_j$ are noise terms, independent of $X_j$ and jointly independent.

We will first show that there is an infinite continuum of options for a structural equation that writes $\bar{Y}$
in terms of $\bar{X}$
 together with an appropriately constructed (formal) noise term. From these options, we will later choose one that renders the causal relation between $\bar{X}$ and $\bar{Y}$ unconfounded. 

To simplify notation, we introduce the auxiliary variable $W:= \sum_{j=1}^N 
f_j(X_j)$ and obtain
\begin{equation}\label{eq:W}
\bar{Y} = W + \bar{V},
\end{equation}
with 
$\bar{V}:= \sum_{j=1}^N V_j$.  

For our coordinate change we can restrict our attention to the two dimensional subspace spanned by $\bar{X},W$: let
$\psi: \R^2\to \R^2$
be a bijection that leaves the first component invariant, that is $\psi(a,b)=(a,\psi_2(a,b))$. We then have
$\psi^{-1}(a,b)=(a,\phi(a,b))$,
where $\phi$ denotes the second component of $\psi^{-1}$. 

We define an additional noise variable $M:= \psi_2(\bar{X},W)$, from which we can reconstruct 
$W$ via $W = \phi (\bar{X},M)$, and rewrite \eqref{eq:W} as 
\begin{equation}\label{eq:nonlinconf}
\bar{Y} = \phi(\bar{X},M) + \bar{V},
\end{equation}
with $\bar{V}$ being independent of $\bar{X}$ and $M$, but $M$ possibly dependent of $\bar{X}$. 
Then,
\eqref{eq:nonlinconf} can be interpreted as
the SCM of a (possibly confounded) causal relation between $\bar{X} $ and $\bar{Y}$ with vector valued noise variable $(M,\bar{V})$. 

To see that $M$ can be chosen in a way that renders this relation unconfounded, define 
$M$ via the conditional cumulative distribution function
$
M(\bar{x},w):= 
P(W\leq w | \bar{X} = \bar{x}). 
$
Whenever the conditional distribution of $W$, given $\bar{X}$ is continuous, $M$ given $\bar{X}$ is uniformly distributed for all $\bar{x}$ and thus independent of $\bar{X}$. The function $\phi$ exists if $M(\bar{x}, \cdot)$ is invertible for all $\bar{x}$. 
With such a choice of $M$,
\eqref{eq:nonlinconf} is the SCM of an unconfounded causal relation between $\bar{X}$ and $\bar{Y}$ and a micro-realization that keeps $M$ constant is then confounding-inhibiting. 

The different choices of $M$ may differ with respect to interpretability and the one above (which renders the relation unconfounded) may not be the most interpretable one. To see this, let us revisit the example $f_j(X_j)=\alpha_j X_j$, where $\alpha_j$ is the price per unit and $V_j:=0$. We can then define the noise $M$ by the average price $
M:= (\sum_{j=1}^N \alpha_j X_j)/(\sum_{j=1}^N X_j)$, 
for which we obtain the SCM $\bar{Y} = \bar{X} \cdot M$.
We can then think of 
an intervention in which the company 
starts selling products to an additional country with 
buying patterns comparable to their existing customers. 
This increases the sales of all price segments by the same percentage, and thus defines an intervention that keeps the average price $M$ constant. Generically, we will have $M \not \indep \bar{X}$, and end up in the confounded scenario with an interpretable noise.

\subsubsection{Macro-interventions for which we cannot specify a macro confounder.}

So far we have shown that the macro variables appear confounded or not, depending on the specified macro interventions. As a somewhat negative result, it turns out there exist macro
interventions which do not make the macro variables
look unconfounded, but also do not allow for an explicit construction of a macro-confounder. Due to space limitation, we elaborate on this in Appendix~\ref{app:macro_confounder}.

\subsection{Macro-confounding in confounded micro-models}\label{sec:confounded} 
Even if the micro systems themselves are confounded, we may still at times get unconfounded macro-models.  In \ref{subsec:4.2}, we first discuss the linear Gaussian setting, and describe conditions for unconfoundedness. In \ref{subsec:4.1} we also present a general technical result for categorical variable models. 

\subsubsection{Linear gaussian confounded models}\label{subsec:4.2}
Assume we are given the confounded linear gaussian model
\begin{equation}\label{eq:conf}
\bar{Y} = \mathbf{\alpha}^T X + N, 
\end{equation}
where $N$ is a noise variable that correlates with $X$ but is not a descendant of $X$.


Let $\pi$ be the aggregation that sums all elements of $X$. Is it possible to define $P^{do}_{\pi, \bar{x}}(X)$ in a way that is confounding inhibiting, that is, 
$P(\bar{Y}|do(\bar{x})) = P(\bar{Y}|\bar{x})$?
The following necessary condition shows that it is not always possible, but we also derive conditions under which it is possible.
Let $\Sigma^{do(\bar{x})}_X$ denote the covariance matrix of $P^{do}_{\pi,\bar{x}}(X)$. Then 
$P(\bar{Y}|do(\bar{x}))$ has variance 
\begin{equation}\label{eq:lingauss_condition} 
{\rm Var}(\bar{Y}|do(\bar{x}))= {\rm Var}(N) + \mathbf{\alpha}^T \Sigma^{do(\bar{x})}_X \mathbf{\alpha}. 
\end{equation} 
Since the second term is non-negative, we can only suppress confounding if 
\begin{equation}\label{eq:varc}
{\rm Var}(\bar{Y}|\bar{x}) \geq  {\rm Var}(N).  
\end{equation} 
We can give \eqref{eq:varc} a geometric interpretation if 
we focus on random variables with zero mean without loss of generality and think of covariance as inner  product in the Hilbert space 
of random variables with finite variance. We can even restrict the attention to 
the 3-dimensional space 
spanned by $\bar{X},\mathbf{\alpha}^T X,N$.

Accordingly, we have ${\rm Var}(N) =\|N\|^2$.
Further, let $Q^\perp$ denote the projection onto the orthogonal complement of $\bar{X}$. Then ${\rm Var}(\bar{Y}|\bar{x})= \|Q^\perp( N + \mathbf{\alpha}^T X)\|^2$ for all $\xbar{x}$, because
the conditional variance is given 
after regressing $\bar{X}$ out and is homoscedastic since we considering joint-Gaussian variables.
We can thus rewrite \eqref{eq:varc} as
\begin{equation}\label{eq:geo} 
 \|Q^\perp( N + \mathbf{\alpha}^T X)\| \geq \|N\|.
\end{equation}

To see that \eqref{eq:varc}
is also sufficient to enable confounding inhibiting interventions, we 
will briefly check that 
$\mathbf{\alpha}^T \Sigma^{do(\bar{x})}_X \mathbf{\alpha}$ can be made arbitrarily large, and then it follows \eqref{eq:varc} is sufficient to guarantee existence of a Gaussian micro-realization that is confounding-inhibiting. Whenever the two vectors
 $\mathbf{\alpha}$ and $\mathbf{1}=(1,\dots,1)^T$ are linearly independent (otherwise the problem is anyway trivial), this is achieved by choosing $\Sigma^{do(\bar{x})}_{X}$ with $\mathbf{1}$ being an eigenvector with eigenvalue $0$ and other eigenvalues large enough so that $\alpha ^T \Sigma^{do(\bar{x})}_X \alpha = \text{Var}(\bar{Y}|\bar{x}) - \text{Var}(N)$.

We now need to show that we can define $do(\bar{x})$
in a way that ensures that conditional
expectations also match:
\begin{equation}\label{eq:exp}
E[\bar{Y}|do(\bar{x})] = E[\bar{Y}|\bar{x}].
\end{equation} 
Since the intervention cannot affect $N$ by assumption,
 \eqref{eq:conf} implies $E[\bar{Y}|do(\bar{x})] = \sum_j \alpha_j E[X_j |do(\bar{x})]  +  E[N]$.

In defining our intervention, we can freely choose 
each  $E[X_j |do(\bar{x})]$ with the only constraint 
 $\sum_j E[X_j |do(\bar{x})] = \bar{x}$. Whenever
 $\mathbf{\alpha}$ and $(1,\dots,1)$ are linearly independent, the term $\sum_j \alpha_j E[X_j |do(\bar{x})] $
 can be made to achieve {\it any} real value, hence we can certainly ensure \eqref{eq:exp}.

 \begin{wrapfigure}{r}{0.5\textwidth}
    \centering
    \resizebox{0.45\textwidth}{!}{%
	\begin{tikzpicture}[roundnode/.style={circle, draw=black!100, very thick, minimum size=0.5mm}, greynode/.style={circle, draw=black!100, fill=black!15, very thick, minimum size=0.5mm}, normalnode/.style={circle, draw=black!0, very thick, minimum size=0.5mm}, squarednode/.style={rectangle, draw=red!60, fill=red!5, very thick, minimum size=5mm}, arrowhead/.style={Triangle[length=3mm, width=2mm]}]
	    \node[roundnode] (z) at (0.5, 3) {$Z$};
		\node[roundnode] (x) at (-0.8,1) {$X$};
		\node[roundnode] (xbar) at (-0.8, -1) {$\bar{X}$};
		\node[roundnode] (y) at (1.8, 1) {$Y$};
		\node[roundnode] (ybar) at (1.8, -1) {$\bar{Y}$};
		
		\draw[-{Triangle[length=3mm, width=2mm]}, line width=1pt] (z) -- (x);
		\draw[{Triangle[open, length=2mm, width=2mm]}-{Triangle[open, length=2mm, width=2mm]}, line width=1pt] (x) -- node[font = {\small}, left=2pt]{$\pi_X$} (xbar);
		\draw[{Triangle[open, length=2mm, width=2mm]}-{Triangle[open, length=2mm, width=2mm]}, line width=1pt] (y) -- node[font = {\small}, right=2pt]{$\pi_Y$} (ybar);
		\draw[-{Triangle[length=3mm, width=2mm]}, line width=1pt] (x) -- (y);
		\draw[-{Triangle[length=3mm, width=2mm]}, line width=1pt] (z) -- (y);
  	    \node[roundnode] (z) at (4.4, 3) {$Z$};
		\node[roundnode] (x) at (3.1,1) {$X$};
		\node[roundnode] (xbar) at (3.1, -1) {$\bar{X}$};
		\node[roundnode] (y) at (5.7, 1) {$Y$};
		\node[roundnode] (ybar) at (5.7, -1) {$\bar{Y}$};

		\draw[-{Triangle[length=3mm, width=2mm]}, line width=1pt] (xbar) --  (x);
		\draw[{Triangle[open, length=2mm, width=2mm]}-{Triangle[open, length=2mm, width=2mm]}, line width=1pt] (y) -- node[font = {\small}, right=2pt]{$\pi_Y$} (ybar);
		\draw[-{Triangle[length=3mm, width=2mm]}, line width=1pt] (x) -- (y);
		\draw[-{Triangle[length=3mm, width=2mm]}, line width=1pt] (z) -- (y);
	\end{tikzpicture}
    }
    \caption{Confounded cause-effect relation between $X$ and $Y$. \textbf{Left}: $G^a$. The amalgamated graph before macro intervention on $\bar{X}$. \textbf{Right}: $G^{a'}$. The amalgamated graph after macro intervention on $\bar{X}$. }
    \label{fig:discrete_micro_confounding}
    \vspace{-10pt}
\end{wrapfigure}
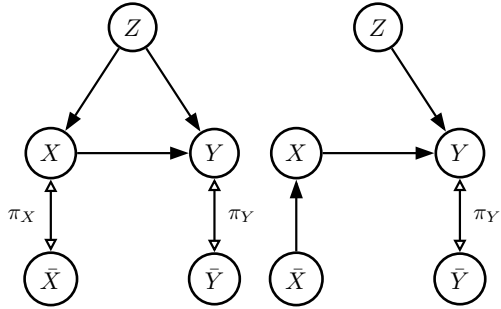

Finally, for which values of $\alpha$ is \eqref{eq:geo} (or equiv., \eqref{eq:varc}) satisfied? When $\alpha \propto \mathbf{1}$, it is clearly not satisfied: since $Q^\perp \bar{X} = 0$ and $\|Q^\perp N\|<\|N\|$ as $\bar{X}$ and $N$ are not perpendicular under the covariance inner product. But as soon as $\alpha$ and $\mathbf{1}$ are linearly independent, the RHS of \eqref{eq:lingauss_condition} can be made arbitrarily large by scaling up $\alpha$ while having it always point in the same direction. Thus, it is possible to tune $\alpha$ as a hyperparameter of the system to ensure that confounding inhibiting interventions exist.

\paragraph{Interpretation.} For the relation between 
sold items $X_i$ of each product $i$ with price $\alpha_i$ and $Y_i =\alpha_i X_i$ its revenue, let $Z_i$ be the number of extra products sold on promotion with reduced price $\gamma_i$. Assume that if the customer buys one item on reduced price, then they will also buy one on full price, giving rise to the micro-level SCM $X_i \defeq Z_i$ and $Y_i \defeq \alpha_i X_i + \gamma_i Z_i$.
Here,  $Z_i$ are micro confounders of the cause-effect pair $X_i\to Y_i$ and we obtain an aggregated confounder
$N:=\sum_i \gamma_i Z_i$, which may disappear
for appropriate interventions on $\bar{X}$, provided the prices $\alpha_i$ are `different enough' from each other.

 The above formulation asserts that the shop owner selling the items can make a confounding-inhibiting micro-realization on the total number of items sold on full price ($\bar{X}$) when full prices $\boldsymbol{\alpha}$ contain enough heterogeneity wrt to the reduced prices $\boldsymbol{\gamma}$ i.e.  When all prices are the same, there is not enough heterogeneity in the micro-model to blur out the confounding. On the other hand, if there is enough heterogeneity in the prices, then the shop owner may find a macro-intervention on the items sold on full prices, for which they can infer the causal effect of, by directly regressing total revenue on total items sold on full price. 
\vspace{-8pt}
\subsubsection{Discrete confounded micro models} \label{subsec:4.1}
\vspace{-2pt}
In the case of nonlinear and non-Gaussian models, we provide at least a result in the case with discrete variables. Our result says that even if the system remains confounded after aggregation in the effect variable, the system may still regain unconfoundedness after aggregation of the cause variables. The proof is in Appendix~\ref{app:discrete}.
\vspace{-5pt}

\begin{thm} \label{thm:1}
    Let $X, Y, Z$ be categorical variables with $|\cX|, |\cY|, |\cZ| < \infty$. Let $G$ be the causal graph with $X\rightarrow Y$, $X\leftarrow Z \rightarrow Y$, and let $\pi_X: \cX \longrightarrow \bar{\cX}$ and $\pi_Y: \cY \longrightarrow \bar{\cY}$ denote aggregation maps. Let $G^a$ be the amalgamated graph of $G$, $\pi_X$ and $\pi_Y$. Further, let ${G^a}'$ be the post-intervention graph for a macro-intervention on $\bar{X}$. See Figure~\ref{fig:discrete_micro_confounding}  Let $P$ denote probability distributions satisfying the conditional independence structure given by $G^a$, and $P^{do}$ denote probability distributions satisfying the independence structure given by ${G^{a}}'$. Suppose $|\bar{\cY}| < \text{min}\left(|\cX|, |\cZ|\right)$, $|\bar{\cX}| < |\cX|$. Then for any coarsening maps $\pi_X$ and $\pi_Y$, and any family of macro interventions on $\bar{X}$: $\{ P^{do}_{\pi_X,\bar{x}}\; |\; \bar{x} \in \bar{\cX}, \; P^{do}_{\pi_X,\bar{x}} \neq P_{X|\bar{X}=\bar{x}}\}$ there exist at least one $P$ such that
\begin{enumerate}
    \item $P(Y|X) \not \equiv P^{do}(Y|X),\; P(\bar{Y}|X) \not \equiv P^{do}(\bar{Y}|X)$. i.e. $(X, Y)$ confounded, $(X, \bar{Y})$ confounded.
    \item $P(\bar{Y}|\bar{X}=
    \bar{x}) = P^{do}(\bar{Y}|\bar{X}=
    \bar{x})$ for all $\bar{x} \in \bar{\cX}$. i.e. $\{(\bar{x}, P^{do}_{X|\bar{X}})\; |\bar{x} \in \bar{\cX}\}$ are confounding-inhibiting micro-realizations.
\end{enumerate}
\end{thm} 
\vspace{-10pt}

\begin{wrapfigure}{r}{0.5\textwidth}
    \centering
\resizebox{0.45\textwidth}{!}{%
	\begin{tikzpicture}[roundnode/.style={circle, draw=black!100, very thick, minimum size=0.5mm}, greynode/.style={circle, draw=black!100, fill=black!15, very thick, minimum size=0.5mm}, normalnode/.style={circle, draw=black!0, very thick, minimum size=0.5mm}, squarednode/.style={rectangle, draw=red!60, fill=red!5, very thick, minimum size=5mm}, arrowhead/.style={Triangle[length=3mm, width=2mm]}]
		\node[roundnode] (x) at (-1,1) {$X$};
		\node[roundnode] (xbar) at (-1, -1) {$\bar{X}$};
		\node[roundnode] (y) at (2, 1) {$Y$};
		\node[roundnode] (ybar) at (2, -1) {$\bar{Y}$};
  		\node[roundnode] (z) at (5, 1) {$Z$};
		\node[roundnode] (zbar) at (5, -1) {$\bar{Z}$};
		\draw[{Triangle[open, length=2mm, width=2mm]}-{Triangle[open, length=2mm, width=2mm]}, line width=1pt] (x) -- node[font = {\small}, left=2pt]{$\pi_X$} (xbar);
		\draw[{Triangle[open, length=2mm, width=2mm]}-{Triangle[open, length=2mm, width=2mm]}, line width=1pt] (y) -- node[font = {\small}, right=2pt]{$\pi_Y$} (ybar);
        \draw[{Triangle[open, length=2mm, width=2mm]}-{Triangle[open, length=2mm, width=2mm]}, line width=1pt] (z) -- node[font = {\small}, right=2pt]{$\pi_Z$} (zbar);
		\draw[-{Triangle[length=3mm, width=2mm]}, line width=1pt] (x) -- (y);
  	\draw[-{Triangle[length=3mm, width=2mm]}, line width=1pt] (y) -- (z);
   \node[roundnode] (xbar) at (-1,-2.5) {$\bar{X}$};
		\node[roundnode] (ybar) at (2, -2.5) {$\bar{Y}$};
   		\node[roundnode] (zbar) at (5, -2.5) {$\bar{Z}$};
		\draw[-{Triangle[length=3mm, width=2mm]}, line width=1pt] (xbar) -- (ybar);
  	     \draw[-{Triangle[length=3mm, width=2mm]}, line width=1pt] 
        (ybar) -- (zbar);
	\end{tikzpicture}
}

    \caption{The {\it natural micro-realization} of macro interventions from Section \ref{subsec:positive} enables coarse graining the  chain (top) to the chain (lower) whenever $\bar{X}\indep \bar{Z}\,|\bar{Y}$.}
    \vspace{-30pt}
    \label{fig:chain}
\end{wrapfigure}
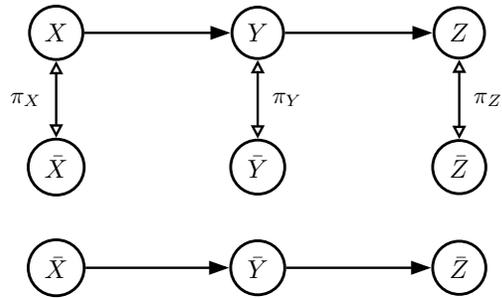

The conditions $|\bar{\cY}| < \text{min}(|\cX|,|\cZ|)$ and $|\bar{\cX}|<|\cX|$ are sufficient conditions which ensure that there is strong-enough aggregation. The theorem thus states that we could find joint distributions over $X, Y, \bar{X}, \bar{Y}$ such that even if both $(X, Y)$ and $(X, \bar{Y})$ are confounded, the confounding can be `cancelled out' for $(\bar{X}, \bar{Y})$ provided there is enough aggregation. 

\vspace{-10pt}
\section{Multi-variate models}\label{sec:general} 
\vspace{-2pt}
In this section we want to follow up on the positive result of
Theorem \ref{thm:main} and 
discuss generalizations and their
challenges. 
\vspace{-10pt}
\subsection{Linear chain} \label{sec:linear-chain}
\vspace{-2pt}


Given the micro model in Figure \ref{fig:chain}, top, we 
are now facing two challenges if we wish to aggregate 
it to the structure on the bottom. First, the macro variable
$\bar{Y}$ will not necessarily block the influence of $\bar{X}$ on $\bar{Z}$ due to information transmission through the micro state $Y$. 
Second, the generalization of {\it natural micro-realization}
raises the following ambiguity for $do(\bar{y})$: to someone who focuses on
the cause-effect relation $Y\to Z$ only, it should be $P(Y|\bar{y})$, aligned with Section \ref{subsec:positive}.
However, after seeing $\bar{X}$ one may want to consider 
$P(Y|\bar{y},\bar{x})$ more natural. This appears to be the right choice at least if one restricts the analysis to the sub-population with fixed $\bar{x}$. Remarkably, these two questions are related: 
\begin{lemma}[irrelevance of cause of causes] \label{lem:chain} 
Given the causal chain in Figure \ref{fig:chain},  then setting $do(\bar{y})\defeq P(Y|\bar{y})$ results in the same
downstream effect on $\bar{Z}$ as implementing it according to  $P(Y|\bar{y},\bar{x})$ if and only if  
\begin{equation}\label{eq:chain} 
\bar{X} \indep \bar{Z} \,|\bar{Y}.   
\end{equation} 
\end{lemma}

If \eqref{eq:chain} does not hold, the chain in Figure \ref{fig:chain}, lower, is not a valid aggregation 
and requires an additional arrow $\bar{X}\to \bar{Z}$ instead.\footnote{In the context of dynamical processes, this question translates to the difficult question of which macro variables are required to describe the relevant part of the history, see e.g. \cite{Crutchfield1999}.}
Whenever \eqref{eq:chain} holds, the cause-effect relation $\bar{Y}\to \bar{Z}$ is valid with respect to both micro-realizations $P(Y|\bar{y},\bar{x})$ 
and $P(Y|\bar{y})$, and they result in same effect on $\bar{Z}$.
Since the natural micro-realization of $do(\bar{x})$ has no ambiguity and renders the relation to $\bar{Y}$ and $\bar{Z}$
unconfounded, the chain in Figure \ref{fig:chain}, lower,
captures all causal relations {\it w.r.t.} natural micro-realizations. Conveniently, \eqref{eq:chain} is a purely statistical criterion about macro variables only.

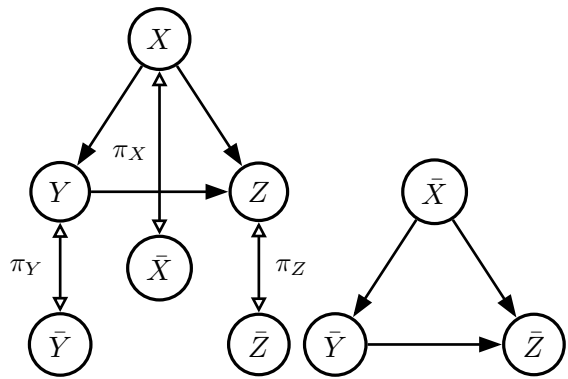
\begin{wrapfigure}{r}{0.5\textwidth}
\resizebox{0.5\textwidth}{!}{%
	\begin{tikzpicture}[roundnode/.style={circle, draw=black!100, very thick, minimum size=0.5mm}, greynode/.style={circle, draw=black!100, fill=black!15, very thick, minimum size=0.5mm}, normalnode/.style={circle, draw=black!0, very thick, minimum size=0.5mm}, squarednode/.style={rectangle, draw=red!60, fill=red!5, very thick, minimum size=5mm}, arrowhead/.style={Triangle[length=3mm, width=2mm]}]
	    \node[roundnode] (x) at (0.5, 3) {$X$};
		\node[roundnode] (y) at (-0.8,1) {$Y$};
		\node[roundnode] (ybar) at (-0.8, -1) {$\bar{Y}$};
		\node[roundnode] (z) at (1.8, 1) {$Z$};
		\node[roundnode] (zbar) at (1.8, -1) {$\bar{Z}$};
        \node[roundnode] (xbar) at (0.5, 0) {$\bar{X}$};  
		
		\draw[-{Triangle[length=3mm, width=2mm]}, line width=1pt] (x) -- (y);
         \draw[-{Triangle[length=3mm, width=2mm]}, line width=1pt] (x) -- (z);
		\draw[{Triangle[open, length=2mm, width=2mm]}-{Triangle[open, length=2mm, width=2mm]}, line width=1pt] (y) -- node[font = {\small}, left=2pt]{$\pi_Y$} (ybar);
		\draw[{Triangle[open, length=2mm, width=2mm]}-{Triangle[open, length=2mm, width=2mm]}, line width=1pt] (z) -- node[font = {\small}, right=2pt]{$\pi_Z$} (zbar);
		\draw[-{Triangle[length=3mm, width=2mm]}, line width=1pt] (y) -- (z);
        \draw[{Triangle[open, length=2mm, width=2mm]}-{Triangle[open, length=2mm, width=2mm]}, line width=1pt] (x) -- node[font = {\small}, left=0pt]{$\pi_X$} (xbar);
        
	    \node[roundnode] (xbar) at (4.1, 1) {$\bar{X}$};
		\node[roundnode] (ybar) at (2.8, -1) {$\bar{Y}$};
		\node[roundnode] (zbar) at (5.4, -1) {$\bar{Z}$};
		\draw[-{Triangle[length=3mm, width=2mm]}, line width=1pt] (xbar) -- (ybar);
         \draw[-{Triangle[length=3mm, width=2mm]}, line width=1pt] (xbar) -- (zbar);
		\draw[-{Triangle[length=3mm, width=2mm]}, line width=1pt] (ybar) -- (zbar);
    	\end{tikzpicture}
     }
    \caption{Left: complete DAG with micro variables $X,Y,Z$, together with its aggregations $\bar{X},\bar{Y},\bar{Z}$. Right: The aggregated DAG which we would like to give a causal semantics by introducing appropriate interventions, if possible.}
    \label{fig:XYZ}
    \vspace{-15pt}
\end{wrapfigure}

\subsection{Backdoor adjustments} \label{sec:backdoor-adjustments}

Assume now we have the DAG in Figure \ref{fig:XYZ}, left.

Under which conditions can we define micro-realizations of macro interventions with respect to which the system behaves 
like the DAG on the right of Figure \ref{fig:XYZ}? We already know that $P(X|\bar{x})$ yields effects on $\bar{Y},\bar{Z}$ that align with the observational conditionals $P(\bar{Y}|\bar{X})$ and $P(\bar{Z}|\bar{X})$, respectively. The questionable part is the effect of interventions on $\bar{Y}$. We need to define them in a way that ensures
\begin{align}\label{eq:backdoor}  
P(\bar{Z}| do(\bar{y}), \bar{x}) =  P(\bar{Z}| \bar{y}, \bar{x}), 
\end{align} 
that is, the backdoor adjustment formula with $\bar{X}$ as adjustment variable (Rule 2, Theorem 3.4.1 in \cite{pearl09}). We find the following sufficient condition:

\begin{lemma}[macro backdoor adjustment]\label{lem:backdoor} 
Given the DAG Figure \ref{fig:XYZ}, left, let $\bar{Z}_y$ denote the random variable
$\bar{Z}$ after adjusting $Y$ to some fixed $y$ 
in the SCM for $\bar{Z}$. 
If 
\begin{equation}\label{eq:back} 
\bar{Z}_y \indep Y \,| \bar{X}, 
\end{equation} 
then 
\eqref{eq:backdoor} holds if $do(\bar{y})$ is defined by
randomizing $Y$ according to $P(Y|\bar{y},\bar{x})$.
\end{lemma}

\vspace{-5pt}
Lemma~\ref{lem:backdoor} states that we can consider the DAG in Figure~\ref{fig:XYZ}, right, as a valid aggregation of the one on the left when we make the macro-intervention $P(Y|\bar{y},\bar{x})$, provided that $\bar{X}$ blocks the backdoor path between 
$Y$ and $\bar{Z}$. Condition \eqref{eq:back} is strong, since it refers to the micro state of $Y$; finding conditions that are easier to handle is left to the future. 
Nevertheless, it is worth mentioning that the proof of Lemma \ref{lem:backdoor} straightforwardly generalizes to back-door adjustments in arbitrary DAGs.
We first state Definition 3.3.1 in 
 \citep{pearl09}.

 \vspace{-3pt}
 \begin{definition}[back-door criterion]
 A set $B$ of variables satisfies the back-door criterion relative to the ordered pair $(X_i,X_j)$ if  (i) no node in $B$ is a descendant of $X_i$ and (ii) $B$ blocks every path between $X_i$ and $X_j$ that contains an arrow into $X_i$.
 \end{definition}

\vspace{-10pt}
\begin{thm}[Generalized backdoor adjustment]\label{thm:backdoor}
Let $G$ be the causal DAG for $X_1,\dots,X_n$
and $P(\bar{X}_1,\dots,\bar{X}_n)$
be Markovian relative to some DAG $\bar{G}$ 
that contains $\bar{X}_i\to \bar{X}_j$ 
if $X_i\to X_j$ is an arrow in $G$, but possibly 
also additional arrows. 
Let the set $B$
of macro variables satisfy the back-door criterion with respect to the pair $(\bar{X}_i,\bar{X}_j)$ in $\bar{G}$ and 
\begin{equation}\label{eq:blockB} 
(\bar{X}_j)_{x_i} \indep 
X_i \,| B,   
\end{equation}
where $(\bar{X}_j)_{x_i}$ denotes 
$\bar{X}_j$ after adjusting $X_i$ to $x_i$.
Then we have 
$
P(\bar{X}_j| do(\bar{x}_i),b) =
P(\bar{X}_j| \bar{x}_i,b)$
with respect to the micro-realization
$P(X_i| \bar{x}_i,b)$.
\end{thm}
\vspace{-5pt}

Whenever there are different adjustment sets
$B,B'$ with respect to which condition \eqref{eq:blockB} holds, the
micro-realizations $P(X_i| \bar{x}_i,b)$ and   $P(X_i|\bar{x}_i,b')$ result in the same effect on $\bar{X}_j$. This is because in any DAG, interventional distributions computed from different back-door adjustments coincide. 

\vspace{-10pt}
\section{Example application scenarios}
Here we detail two scenarios where our framework can be used in practice.

\paragraph{Unconfounded micro-model.} A large retail company has millions of products for sale. The micro variables are sales of products in some same category $X_1 ,…, X_K$, for example, $X_i$ is the number of bottles sold for the $i^{th}$ type of shampoo, and product revenues $Y_1,…,Y_K$. The value of K is large, i.e. in the thousands. The micro-variable structural equations are $Y_i = \alpha_i X_i + N_i$, $X_i = \mathcal{N}(\mu_i, \sigma_i)$, and $N_i \perp X_i$. For the management, a macro model is presented, where the macro variables are $\bar{X}$ and $\bar{Y}$. The management would like to enable easy causal inference at the macro level by reading off the causal effect of changing $\bar{X}$ on $\bar{Y}$ from the observational distribution $P(\bar{Y} | \bar{X})$. This instructs the natural micro-realization, that is $P^{do}_{\pi}(X|\bar{X}=\bar{x}) := P(X|\bar{X}=\bar{x})$ when implementing the macro intervention $\bar{X} := \bar{x}$. Note, that the natural micro realization does not reference the micro level structural equations. In practice, the natural implementation here is a Multivariate Gaussian density whose parameters are functions of $\bar{x}$, so the company can sample from the density and set $X_i,…,X_K$ to the sampled value. The caveat is that in practice this procedure can become costly.

\paragraph{Confounded micro-model.} Now imagine the same micro-level model, but this time $X_i$ is not independent of $N_i$. There are two ways to proceed. (i) The management still wants to read off causal effect directly from the observational distribution $P(\bar{Y}|\bar{X})$. Then the company needs to implement a confounding inhibiting micro realization, for which we described the derivation and sufficient conditions in section 3.2.1. Note that the sufficient condition, i.e. having enough variability in the prices $\alpha_i$ now depends on the micro structural equations. (ii) If we can define $\bar{N}$ as a function of $X_1,…,X_K, N_1,…,N_K$ such that equation \eqref{eq:back} is satisfied, that is $\bar{Y}_x \perp X | \bar{N}$ then the company can also perform a macro-backdoor adjustment $P(\bar{Y}| do(\bar{X})) = \sum_{\bar{N}}P(\bar{Y} | \bar{X}, \bar{N}) P(\bar{N})$.

\vspace{-5pt}
\section{Conclusion}\label{sec:conclusion}
\vspace{-5pt}
Aggregated variables are the rule rather than the exception 
in causal models of processes in everyday life. We highlight three challenges for causal aggregation which are not sufficiently addressed by existing work: 1) causal aggregation induces ambiguity, 2) bridging macro- and micro-interventions, and 3) the micro-graph may not be known. This paper intensifies Challenge 1 and shows that causal aggregation entails 
ambiguity not only with respect to the quantitative effect, but even with respect to the causal structure, namely whether a cause-effect relation is confounded or not. On the positive side, this paper addresses Challenges 3 by introducing `natural micro-realizations', defining which does not require knowledge of the causal graph. Moreover, in introducing `natural macro-interventions' and `confounding-inhibiting (inducing) micro-realizations', we naturally address Challenge 2 and open up the way to ask questions of `What are the best micro-realizations for a given macro-intervention under a given goal such as no macro-confounding?'. 

\acks{We gratefully acknowledge, in alphabetical order, Sergio Hernan Garrido Mejia, Atalanti Mastakouri and Leena Chennuru Vankadara for helpful discussions and for proofreading the paper. \\
YZ acknowledges support by the Engineering and Physical Sciences Research Council with grant number EP/S021566/1.}

\bibliography{references}

\appendix

\onecolumn 

\appendix
\section{Example from \citet{Rubenstein2017}}\label{sec:rubenstein-example}

\begin{customex}{1}\label{ex:ruben}
Let $M = (\mathbf{S}, P_{\mathbf{N}})$ be a micro causal model where 
\begin{align*}
   \mathbf{S} &= \left\{X_1 \defeq U_1, \; X_2 \defeq U_2, \; Y_1 \defeq \alpha_1 X_1, \; Y_2 \defeq \alpha_2 X_2\right\} 
\end{align*}
where $\alpha_1 \neq \alpha_2$, $\cU_1 = \{0, 1\}$, $\cU_2 = \{0,1\}$, $\text{supp}[P_{U_1}] = \text{supp}[P_{U_2}] = \{0,1\}$. Now, consider a set of interventions that we are interested in on the micro causal model.
\begin{align*}
   \cI = \{&i_1 = do(X_1\defeq 1, X_2 \defeq 0), \\&i_2 = do(X_1 \defeq 0, X_2 \defeq 1)\}.
\end{align*}
as well as aggregation maps $\bar{X}=\pi_X((X_1, X_2)) = X_1 + X_2$ and $\bar{Y} = \pi_Y((Y_1, Y_2)) = Y_1 + Y_2$. 

Now applying $i_1$ to the system results in $X_1=1$, $X_2=0$, $Y_1=\alpha_1$, $Y_2=0$, and thus $\bar{X}=1$ and $\bar{Y}=\alpha_1$. Meanwhile, applying $i_2$ to the system results in $X_1 = 0$, $X_2 = 1$, $Y_1 = 0$, $Y_2 = \alpha_2$, and thus $\bar{X} = 1$ and $\bar{Y} = \alpha_2$. Thus, there is no structural assignment $f: \bar{\cX} \rightarrow \bar{\cY}$ s.t. 
\begin{align*}
   \pi_{\#} P^{do(i_k)} ( X_1, X_2, Y_1, Y_2) &= P^{do(\bar{X} \defeq 1)} (\bar{X}, \bar{Y})
\end{align*}
where $k = 1, 2$.

Therefore, it is not possible to define a sensible causal model on the macro variables: since $X_1, X_2$ precede $Y_1, Y_2$ in causal order, $\bar{X}$ should also precede $\bar{Y}$ if there is a `macro causal model', yet there is no way to define a structural equation from $\bar{X}$ to $\bar{Y}$ that makes the `macro causal model' consistent with the micro model in the sense that the pushforward measure of interventions on the micro model is always equal to the interventional distribution on the macro model given by the corresponding intervention. Here, corresponding intervention is given by the aggregation map. The above intuition can be formalized - \citep{Rubenstein2017} provides a formal definition of consistency, which is violated by our example exactly for the reason we laid out. In fact, only when $\alpha_1 = \alpha_2$ are we able to obtain a consistent macro causal model under the definition of theirs. 
\end{customex}

\section{Digression on Shift Interventions}\label{app:shift_intervention}

The natural micro-realization is a stochastic intervention on the micro-variables. How to implement this is not immediately obvious. Thus, here we suggest shift interventions, which change the value of micro-variables {\it relative} to its current state.
One choice is to consider the simple shift-intervention $X \mapsto X+\delta$ with constant $\delta$, which is a special case of general shift-interventions $X \mapsto f(X)$ \citep{rothenhaeusler:2015:shift-intervention, pmlr-v124-sani20a}. In the standard setting (i.e. without dealing with aggregation), a shift intervention is implemented as the practitioner simply observe whichever treatment the subject is about to receive, and then add a constant on top of it. Crucially, this is a deterministic action. We now elaborate on its relations with atomic interventions. 

When treatments take real values, atomic interventions can be generated from shift-interventions:
\begin{align}
    &\mathbb{P}(Y|do(X \defeq x)) \nonumber\\
    = &\int_{x' \in \R} \mathbb{P}(Y|X=x', do\left(X\defeq X + (x-x')\right))  \nonumber\\
    &\cdot \mathbb{P}(X = x')dx'\\
    = &\int_{x' \in \R} \mathbb{P}(Y|X=x', do(X\defeq x)) \cdot \mathbb{P}(X = x') dx'
\end{align}
and vice versa:
\begin{align}
    &\mathbb{P}(Y|do(X \defeq X + \delta)) \nonumber\\
    = &\int_{x' \in \R} \mathbb{P}(Y|X=x', do\left(X\defeq x'+\delta)\right))  \nonumber\\
    &\cdot \mathbb{P}(X = x')dx'
\end{align}
In other words, we can always generate the effect of an atomic intervention from shift interventions, and vice versa: imagine a practitioner observing a particular treatment-assignment group $(x')$, they then make the appropriate shift-intervention $(x-x')$ to change the treatment to $x$; they do this for every observed $x'$-treatment-assignment group. Then on average, this is the same as doing an atomic intervention to $x$ for the whole group.

We can come up with an equivalent notion of no confounding using shift-interventions. Therefore, we may reason safely with shift-interventions and assured that whenever we conclude unconfoundedness under shift-interventions, we will also conclude unconfoundedness under atomic interventions.
\begin{definition}[Unconfoundedness under $\delta$-shift-interventions for 1-d Linear Gaussian models]
Suppose the causal order is $X \rightarrow Y$. Define the counterfactual variable after a $\delta$-shift intervention as $X^{\delta} \defeq X + \delta$. $X$ is unconfounded with $Y$ if 
\vspace{-1ex}
\begin{align}
    \forall \delta \in \R, \forall x \in \R & \nonumber\\
    \mathbb{P}(Y|X^{\delta} = x) &= \mathbb{P}(Y|X= x)
\end{align}
\end{definition}
\vspace{-1ex}
The intuition of this is the following: imagine patient A and patient B would have received treatment $1.0$ and treatment $1.1$ without intervention. The shift-intervention would add $+0.1$ to all treatments which the patients would have received. So, after the intervention, patient A and patient B would receive treatment $1.1$ and treatment $1.2$. Note that patient $A$ \textit{after} the intervention and patient B \textit{before} the intervention receive the same treatment ($1.1$). If unconfoundedness holds, then patient B before intervention and patient A after intervention would react the same way. 

We now prove the equivalence of this definition to the one using atomic interventions, in the case of linear structural equations.

\begin{customlemma}{4}\label{lemma:shift_atomic_equiv}
Define the post-shift-intervention variable as $X^{\delta} := X + \delta$. For a linear structural equation model, $Y := a X + N$ where $a$ is a constant coefficient, the following two statements are equivalent:
\begin{enumerate}
    \item $\forall x \in \cX = \R, \; P(Y|do(X\defeq x))=P(Y|X=x)$.
    \item $\forall \delta \in \R, \; \forall x \in \R,\; P(Y|X^{\delta}=x)=P(Y|X=x)$, 
\end{enumerate}
\end{customlemma}
\begin{proof}
($2 \implies 1$.)
By definition,
\begin{align}
    P(Y|X^{\delta} = x) &= P(Y|X=x-\delta, do(X\defeq x)) \;\; \forall \delta
\end{align}
By 2, 
\begin{align}
    P(Y|X=x', do(X\defeq x)) = P(Y|X=x) \forall x' \label{eq:shift_cond}
\end{align}
Therefore,
\begin{align}
    P(Y|do(X\defeq x)) &= \int_{\cX} P(Y|X=x', do(X\defeq x)) \underbrace{P(X=x'|do(X=x))}_{=P(X=x')}dx'\\
    &= \int_{\cX} P(Y|X=x)P(X=x')dx' \hspace{5cm} \text{by \eqref{eq:shift_cond}}\\
    &= P(Y|X=x)
\end{align}
($1 \implies 2.$)
By 1,
\begin{align}
    P(Y|do(X\defeq x)) &= \int_{\cX} P(Y|X=x', do(X\defeq x))P(X=x'|do(X\defeq x))dx'\\
    &= \int_{\cX} P(Y|X=x', do(X\defeq x)) P(X=x') dx' \label{eq:atomic_cond1}\\
    &= P(Y|X=x) \;\; \forall x \label{eq:atomic_cond2}
\end{align}
Since $Y=aX + N$, we have
\begin{align}
   \eqref{eq:atomic_cond1} &= \int_{\cX} P(ax+ N|X=x') P(X=x')dx'\\
   &= P(ax+N) \;\;\text{(by basic rules of probability)}\\
   &= P(ax + N|X=x) = \eqref{eq:atomic_cond2} \;\; \forall x 
\end{align}
This means that $N \indep X$, and therefore 1 holds.
\end{proof}

\section{A perspective from linear change of coordinates} \label{app:change-of-cood}
\vspace{-1ex}
The above example can also be formalized by thinking of the vector of micro variables $X = (X_1,\cdots, X_N)$ as a coordinate system and the aggregation as a result of a linear change of coordinates. Let $H: \R^N \rightarrow \R^N$ be a bijective linear map (i.e. the change-of-coordinate map).  Further, requiring the first row of $H$ to be ones would give us the sum aggregation: 
\begin{equation}
    H_{1,:}=(1,1,\dots,1)\label{eq:H_1j} 
\end{equation}
Then, trivially it follows that
\begin{align*}
    \bar{Y} &\defeq \boldsymbol{\alpha}^\top X
    =\boldsymbol{\alpha}^\top H^{-1} H X
    = \underbrace{\sum_{i=1}^N \alpha_i H^{-1}_{i,1}}_{\beta} \bar{X} + \underbrace{\sum_{i, k=1}^N\sum_{j=2}^N\alpha_{i}H^{-1}_{i, j}H_{j,k}X_k}_{\bar{U}_{\beta}} \label{eq:Ybar_new_coord}
\end{align*}
Therefore, after a linear change of coordinates with $\bar{X}$ being part of the new coordinate system, $\bar{Y}$ still admits a linear structural equation under the new coordinates. We can view this as a usual structural equation of $\bar{Y}$, since all we did was changing the coordinates. Now, an intervention on $\bar{X}$ amounts to changing $\bar{X}$ while not affecting remaining coordinates $\sum_{k=1}^N H_{2:, k}X_k$, and hence the noise term, $\bar{U}_{\beta}$, which is a linear combination thereof. 
This translates the ambiguity of $do(\bar{x})$ into specifying 
the basis vectors corresponding to the {\it remaining} coordinates. 

One may ask, since $H$ is constrained - it must satisfy \eqref{eq:H_1j} - it needs to be shown that there exist $H$ so that a given $\beta$ can be obtained. With a little bit more effort, we can also show that for any $\boldsymbol{\Delta}$ such that $\sum_{i}\Delta_i = 1$, there also exist $H$ such that $\boldsymbol{\Delta} = H^{-1}_{:, 1}$. Indeed, we check this in the following lemma. 

\begin{lemma}\label{lemma:beta_obtain}
For a given $\boldsymbol{\alpha} \in \R^N, \; \boldsymbol{\alpha}\not \propto (1,\cdots, 1)$ and $c \in \R$, there exists an invertible $N\times N$ matrix $H \in \R^{N\times N}$ such that $H_{1,i} = 1 \; \forall i=1,\cdots, N$ and $\sum_{i}\alpha_i H^{-1}_{i, 1} = c$. Moreover, for any $\boldsymbol{\Delta}$ such that $\sum_{i}\Delta_i = 1$, there also exist $H$ such that $\boldsymbol{\Delta} = H^{-1}_{:, 1}$.
\end{lemma}

The proof is in Appendix~E.
Lemma~\ref{lemma:beta_obtain} shows that for any $\beta$, there exist at least one corresponding change-of-coordinate matrix $H$. This is remarkable: When $\beta = \frac{\text{Cov}(\bar{Y}\bar{X})}{\text{Cov}(\bar{X}, \bar{X})}$, we know from regression that $\bar{X}$ and $\bar{Y}$ look unconfounded. On the contrary, note that $\bar{U}_{\beta} = \boldsymbol{\alpha}^\top X - \beta \bar{X}$. Therefore, as $\beta \rightarrow \infty$, $|{\rm Cov}(\bar{X}, \bar{U}_{\beta})| \rightarrow \infty$. This means that depending on the change-of-coordinate matrix chosen, $\bar{X}$ can look either unconfounded or \textit{arbitrarily strongly confounded with $\bar{Y}$}. 

Relating the ambiguity of macro interventions to the ambiguity of the coordinate systems links this work also to causal representation learning, which deals with causal modeling in scenarios where the variables are not given a priori \citep{Bengio2013,Schoelkopf2021}.  
\subsection{Shift interventions under a change-of-coordinate.} \label{app:shift-change-of-coord}
Clearly, the change-of-coordinate perspective also implies, as a consequence, how to distribute the shift in $\bar{X}$ to shifts in the micro-variables $X_1,\cdots, X_N$. From \eqref{eq:Ybar_new_coord}, doing $\bar{X} \mapsto \bar{X} + 1$ amounts to doing $X \mapsto X + H^{-1}_{:, 1}$:
\begin{align}
    \bar{Y} (\bar{X} + 1) &= \underbrace{\boldsymbol{\alpha}^\top H^{-1}_{:, 1} \bar{X} + U_\beta}_{\boldsymbol{\alpha}^\top X} + \boldsymbol{\alpha}^\top H^{-1}_{:, 1}
    = \boldsymbol{\alpha}^\top (X + H^{-1}_{:, 1})
\end{align}
i.e. shifting by $1$ on $\bar{X}$ and shifting $X$ by $H^{-1}_{:, 1}$ on the micro-variables are the same thing.

This makes calculating $\beta$ easy 
if you have already decided how you want to shift $X$: by Lemma~\ref{lemma:beta_obtain}, $H^{-1}_{:, 1}$ can be any vector $\boldsymbol{\Delta}$ whose elements sum to $1$. Therefore, just make sure 
that the coefficients of $\Delta$ 
sum up to $1$, and the corresponding structural coefficient reads $\beta = \boldsymbol{\alpha}^\top \Delta$. Importantly, we realize that by setting $\Delta = \frac{1}{\sum_n \text{Var}[X_n]}(\text{Var}[X_1],\cdots, \text{Var}[X_N])$, we actually obtain the same $\beta$ as that obtained by the natural intervention.

\section{There are macro-interventions for which we cannot specify a macro confounder.}\label{app:macro_confounder}
We have shown that depending on the specified macro interventions, the macro variables appear confounded or not. As we will show next, there also exist macro interventions which do not allow for any model on the macro variables that explains both the observational and interventional distribution. Suppose in the scenario of Section~\ref{subsec:aggre} we have a third shop owner who made the deterministic intervention 
\begin{align}
    \begin{pmatrix}X_1 \\ X_2\end{pmatrix}\Bigg|\bar{X} &\sim \cN \left(\begin{pmatrix}\bar{X} \\ 0\end{pmatrix}, \mathbf{0}\right)
\end{align}
This would induce a deterministic post-intervention relationship between $\bar{\cX}$ and $\bar{\cY}$:
\begin{align}
    \bar{Y}= \alpha_1 \bar{X}
\end{align}
Had there been a structural causal model from $\bar{X}$ to $\bar{Y}$ under this intervention, it would have the form:
\begin{align}
    \bar{Y} &= f(\bar{X}, N)
\end{align}
but since intervening on $\bar{X}$ to set $\bar{Y}$ to a deterministic value, $f$ must be constant in $N$ for every $\bar{X}=\bar{x}$, so wlog we can rewrite the structural equation as 
\begin{align}
    \bar{Y} &= g(\bar{X})
\end{align}
But this implies that the observational random variable, $\bar{Y} | \bar{X}$, is also deterministic, which contradicts our hypothesis wherever $\alpha_1 \neq \alpha_2$. Thus, there may be macro-interventions for which we cannot write down a macro causal model which is possibly confounded. We leave it for future work to further investigate the range of macro-interventions which admit macro-confounding variables.
\section{Proofs}\label{app:proofs}
\subsection{Section~\ref{sec:unconfounded}}

\noindent{\bf Proof of Theorem~\ref{thm:main}.}
To keep notation simple, we will consider discrete variables but the generalization to continuous variables is obvious.

$P(\bar{Y}|do(\bar{x}))= \sum_{x,y} 
P(\bar{Y}|y)P(y|x)P(x|\bar{x})= \sum_x 
P(\bar{Y}|x)P(x|\bar{x})= P(\bar{Y}|\bar{x})$. \hfill\BlackBox

\subsection{Section~\ref{sec:confounded}. Micro-confounding - discrete case}\label{app:discrete}

When in the discrete setting, we use the standard way to construct probability spaces i.e. take the entire set as sample space and its power set as the $\sigma$-algebra.

We outline here the moral reason that the statement should be true. 

For the observational and interventional distributions of $\bar{Y}|\bar{X}$ to be the same, i.e. unconfoundedness in the macro variables, we can consider the decomposition of $P_{\bar{Y}|\bar{X}}$ and $P^{do}_{\bar{Y}|\bar{X}}$. Since the macro interventions, $(\;\bar{x}, P_{X|\;\bar{x}})$ for various values of $\bar{x}$, only impact the generative process for $X$ and $\bar{X}$, we can observe that all other generative mechanisms are invariant before and after the intervention.  Writing down the decomposition, we get
\begin{align}
    P^{do}_{\bar{Y}|\bar{X}}(y) &= \sum_{i,j} P_{\bar{Y}|x_i,z_j}(y) P^{do}_{X|\bar{X}}(x_i) P_Z(z_j) \label{eq:confound1}\\
    P_{\bar{Y}|\bar{X}}(y) &= \sum_{i,j} P_{\bar{Y}|x_i,z_j}(y) P_{X, Z|\bar{X}}(x_i, z_j) \label{eq:confound2}
\end{align}
Unconfoundedness between $\bar{X}$ and $\bar{Y}$ thus amounts to saying that the right-hand-sides of \eqref{eq:confound1} and \eqref{eq:confound2} are the same for all values of $y$. \eqref{eq:confound1} and \eqref{eq:confound2} can be written as vectorised equations, and equality of the right-hand-side amounts to saying that, roughly speaking, the difference between $P^{do}_{X|\bar{X}} \otimes P_Z$ and $P_{X, Z|\bar{X}}$ lie in the null space of $P_{\bar{Y}|X,Z}$. Why should this be true? Morally, this is because the null space of $P_{\bar{Y}|X,Z}$ is large due to the coarsening, thus possible to contain the said difference sometimes. Additional technical conditions need to be satisfied in the theorem, but these can likewise be satisfied by constructing $P_{\bar{Y}|X,Z}$ with the correct null space.
\vspace{+1ex}


\noindent{\bf Proof of Theorem~\ref{thm:1}.}
The set of probability measures with conditional dependence structure encoded by $G^a$ is 
\begin{align}
\mathcal{P}^{obs} &\defeq \bigg\{P \;s.t. \;
    P(X,Z,\bar{X}, \bar{Y}) \equiv P(Z)P(X|Z)P(\bar{X}|X)P(\bar{Y}|X,Z)
    \bigg\}
\end{align}
Meanwhile, $\mathcal{P}^{do}$ denotes the set of probability measures $P^{do}$ on the same sample space and $\sigma$-algebra, but compatible with performing a macro-intervention on $\bar{X}$:
\begin{align}\mathcal{P}^{do} &\defeq \bigg\{P^{do} \; s.t. \;
    P^{do}(X,Z,\bar{X}, \bar{Y}) \equiv P^{do}(Z)P^{do}(\bar{X})P^{do}(X|\bar{X})P^{do}(\bar{Y}|X,Z) \bigg\}
\end{align}
Since the generative processes of all variables except for $\bar{X}$ and $X$ are invariant before and after macro-intervention on $\bar{X}$, write
\begin{align}
    P(Z) &= P(Z) = P^{do}(Z) \\
    P(Y|X,Z) &= P(Y|X,Z) = P^{do}(Y|X,Z)
\end{align}
Therefore, $P(\bar{Y}|X,Z) = P^{do}(\bar{Y}|X,Z)$. So write:
\begin{align}
    P(\bar{Y}|X,Z) &= P(\bar{Y}|X,Z) = P^{do}(\bar{Y}|X,Z)
\end{align}
Note the decompositions:
\begin{align}
    P(Y|X) &= \sum_{j}^{|\cZ|} P(Y|X,Z=z_j) P(Z=z_j|X)\\
    P^{do}(Y|X) &= \sum_{j}^{|\cZ|} P(Y|X,Z=z_j) P^{do}(Z=z_j)\\
    P(\bar{Y}|X) &= \sum_{j}^{|\cZ|} P(\bar{Y}|X,Z=z_j) P(Z=z_j|X)\\
    P^{do}(\bar{Y}|X) &= \sum_{j}^{|\cZ|} P(\bar{Y}|X,Z=z_j) P^{do}(Z=z_j)\\
    P(\bar{Y}|\bar{X}) &= \sum_{i,j}^{|\cX|, |\cZ|} P(\bar{Y}|X=x_i,Z=z_j) P(X=x_i,Z=z_j|\bar{X})\\
    P^{do}(\bar{Y}|\bar{X}) &= \sum_{i,j}^{|\cX|, |\cZ|} P(\bar{Y}|X=x_i,Z=z_j) P^{do}(X=x_i,Z=z_j|\bar{X})\\
\end{align}
Define the following linear maps:
\begin{align}
    f_{1,x}: \; \mathbb{R}^{|\cZ|} &\longrightarrow \mathbb{R}^{|\cY|} \\
    \mathbf{v} &\longmapsto \sum_{j}^{|\cZ|} P(Y|X=x,Z=z_j) v_j\\
    f_{2,x}: \; \mathbb{R}^{|\cZ|} &\longrightarrow \mathbb{R}^{|\bar{\cY}|} \\
    \mathbf{v} &\longmapsto \sum_{j}^{|\cZ|} P(\bar{Y}|X=x,Z=z_j) v_j\\
    f_3: \; \mathbb{R}^{|\cX|\times |\cZ|} &\longrightarrow \mathbb{R}^{|\bar{\cY}|} \\
    M &\longmapsto \sum_{i,j}^{|\cX|, |\cZ|} P(\bar{Y}|X=x_i,Z=z_j)M_{ij}
\end{align}

Then conditions 1, 2, 3 are equivalent to 
\begin{align}
    \exists x \in \cX, \forall \bar{x} \in \bar{\cX}:&\\
    &P(Z|X=x) - P^{do}(Z) \not\in Ker( f_{1,x} ) \label{eq:34}\\
    &P(Z|X=x) - P^{do}(Z) \not\in Ker( f_{2,x} ) \label{eq:35}\\
    &P(X,Z|\bar{X}=\bar{x}) - P^{do}(X,Z|\bar{X}=\bar{x}) \in Ker( f_3 ) \label{eq:36}
\end{align}

Now analyse $P(X,Z|\bar{X})$ and $P^{do}(X,Z|\bar{X})$.
\begin{align}
    P(X,Z|\bar{X}) &= \frac{P(\bar{X}|X)P(X|Z)P(Z)}{\sum_{i,j}^{|\cX|, |\cZ|}P(Z=z_j)P(X=x_i|Z=z_j)P(\bar{X}|X=x_i)}\\
    P^{do}(X,Z|\bar{X}) &= P(Z)P^{do}(X|\bar{X})
\end{align}

\textit{An aside: A probability distribution of a discrete random variable, say $P(A), A \in \cA, |\cA|<\infty$ can be viewed as a finite-dimensional vector $\mathbf{v}$ such that $v_i = P(A=a_i)$. From now on we refer to this as the \textit{vector of $P(A)$}.}

Note that for a given $\bar{x}$, $P(\bar{X}=\bar{x}|X=x)$ and $P^{do}(X=x|\bar{X}=\bar{x})$ must be zero when $x \not\in \pi_X^{-1}(\bar{x})$. Since $P(\bar{X}|X)$ is a factor of $P(X,Z|\bar{X})$, and $P^{do}(X|\bar{X})$ is a factor of $P^{do}(X,Z|\bar{X})$, we can work out the elements of the vectors $P(X,Z|\bar{X}=\bar{x})$ and $P^{do}(X,Z|\bar{X}=\bar{x})$ which are allowed to be non-zero i.e. precisely the elements corresponding to $x$ with $x \in \pi_X^{-1}(\bar{x})$. 
Note that $\pi^{-1}(\bar{x}) \cap \pi^{-1}(\bar{x}') = \emptyset$ when $\bar{x} \neq \bar{x}'$. 

Order the elements of $\bar{\cX}$ as $\{\bar{x}_1,\cdots, \bar{x}_K, \; K=|\bar{\cX}|\}$. Since $|\bar{\cX}| < |\cX|$, we can choose $\bar{x}$ such that $|\pi^{-1}(\bar{x})| > 1$. Wlog, let this be $\bar{x}_1$. Let $P(X^+,Z|\bar{X}=\bar{x}_1)$ and $P^{do}(X^{+, 1},Z|\bar{X}=\bar{x}_1)$ be matrices which contain the elements of $P(X,Z|\bar{X}=\bar{x}_1)$ and $P^{do}(X,Z|\bar{X}=\bar{x}_1)$ such that $X \in \pi^{-1}(\bar{x}_1)$. Precisely, $X^{+, 1}$ is a vector of length $|\pi^{-1}(\bar{x}_1)|$ where $x^{+, 1}_i \in \pi^{-1}(\bar{x}_1) \; \forall i=1,\cdots, |\pi^{-1}(\bar{x}_1)|$. The $ij^{th}$ element of $P(X^{+, 1},Z|\bar{X}=\bar{x}_1)$ is given by $P(X = x^{+, 1}_i, Z=z_j|\bar{X}=\bar{x}_1)$. Choose $P(Z)\text{ and } P(X|Z)$, such that there exist some values of $x^{+, 1}$ such that $P(X=x^{+, 1}|\bar{X}=\bar{x}) \neq P^{do}(X=x^{+, 1}|\bar{X}=\bar{x})$ and $P(Z|X=x^{+, 1}) \neq P(Z)$. Wlog, let this be $x_1^{+,1}$. Then $P(Z|X=x_1^{+, 1}) \text{ and } P(Z)$ are linearly independent since they are both normalised. It follows that $\mathbf{u}_{x_1^{+, 1}}\defeq P(Z|X^{+, 1}=x_1^{+, 1}) - P(Z)$ is linearly independent of $\mathbf{v}_{x_1^{+, 1}} \defeq P(Z|X=x_1^{+, 1})P(X=x_1^{+, 1}|\bar{X}=\bar{x}_1) - P(Z)P^{do}(X=x_1^{+, 1}|\bar{x}_1)$. By Lemma~\ref{lemma:3}, there exist (a distribution which we call) $P(\bar{Y}|X= x_1^{+, 1}, Z)$ which maps $\mathbf{v}_{x_1^{+,1}}$ to $0$ and maps $\mathbf{u}_{x_1^{+, 1}}$'s to a non-zero vector. For the other values of $x_i^{+,k}\in \pi^{-1}(\bar{x}_k),\;\; i, k\neq 1$, we only need to satisfy \eqref{eq:36}. Lemma~\ref{lemma:3} also immediately implies that there is $P(\bar{Y}|X=x_i^{+, k}, Z)$ such that $\mathbf{v}_{x_i^{+, k}} \defeq P(Z|X=x_i^{+, k})P(X=x_i^{+, k}|\bar{X}=\bar{x}_k) - P(Z)P^{do}(X=x_i^{+, k}|\bar{X} = \bar{x}_k)$ is mapped to $0$. Therefore, condition 3 and the second half of condition 1 are satisfied.

It remains to satisfy the first half of condition 1. This is easy, because the linear map $f_{2,x}$ is given by composing $f_{1,x}$ and $\mathbf{v} \in \mathbb{R}^{|\cY|}\mapsto \sum_{k}^{|\cY|} P(\bar{Y}|Y=y_k) v_k$. For a fixed $f_{2,x}$ and $P(\bar{Y}|Y)$, $\exists f_{1,x}$ s.t. $f_{2,x} = P(\bar{Y}|X) \circ f_{1,x}$, since, for example, take $P(Y = y_i|X,Z) = \frac{1}{|\pi^{-1}(\bar{y})|}P(\bar{Y}=\bar{y}|X,Z)$ for any $y_i \in \pi^{-1}(\bar{y})$. Then since $f_{2,x^+}$ maps $\mathbf{u}_{x^+}$ to a non-zero vector, so must $f_{1,x^+}$.






\hfill\BlackBox

\subsection{Section~\ref{sec:linear-chain}}
\noindent{\bf Proof of Lemma~\ref{lem:chain}.}
Implementing $do(\bar{y})$ according to $P(Y|\bar{y})$ results in the downstream effect
$\sum_y P(\bar{Z}|y)P(y|\bar{y})= \sum_y P(\bar{Z}|y,\bar{y})P(y|\bar{y}) = P(\bar{Z}|\bar{y})$. Micro-realization via $P(Y|\bar{y},\bar{x})$ results in $\sum_y P(\bar{Z}|y)P(y|\bar{y},\bar{x}) =\sum_y P(\bar{Z}|y, \bar{y},\bar{x})P(y|\bar{y},\bar{x})= P(\bar{Z}|\bar{y},\bar{x})$.  
The statement $P(\bar{Z}|\bar{y}) =  P(\bar{Z}|\bar{y},\bar{x})$ holds for all $\bar{x},\bar{y}$ iff. \eqref{eq:chain}  is true. 
\hfill\BlackBox

\subsection{Section~\ref{sec:backdoor-adjustments}}
\noindent{\bf Proof of Lemma~\ref{lem:backdoor}.}
For fixed $\bar{x}$, \eqref{eq:back} states that $\bar{Z}$ 
is a function of noise that is independent of
$Y$, thus we have equality of the
interventional and observational probabilities 
$
P(\bar{Z}|do(y),\bar{x}) = P(\bar{Z}|y,\bar{x}).    
$
We conclude 
$
P(\bar{Z}|do(\bar{y}),\bar{x}) = \sum_y  P(\bar{Z}|do(y),\bar{x}) p(y|\bar{y},\bar{x})
= \sum_y P(\bar{Z}|y,\bar{x}) p(y|\bar{y},\bar{x}) = P(\bar{Z}|\bar{y},\bar{x}). 
$
\hfill\BlackBox

\noindent{\bf Proof of Theorem~\ref{thm:backdoor}.}
Immediate by replacing 
$Y$ with $X_i$, $Z$ with $X_j$, and 
$X$ with $B$ in the proof of Lemma \ref{lem:backdoor} (note that the back-door
condition for $\bar{G}$ implies the one for $G$
because there the former can only differ by additional arrows.   \hfill \BlackBox 

\section{Auxiliary lemmas}\label{app:aux}

\begin{customlemma}{1}
For a given $\boldsymbol{\alpha} \in \R^N, \; \boldsymbol{\alpha}\not \propto (1,\cdots, 1)$ and $c \in \R$, there exist an invertible $N\times N$ matrix $H \in \R^{N\times N}$ such that $H_{1,i} = 1 \; \forall i=1,\cdots, N$ and $\sum_{i}\alpha_i H^{-1}_{i, 1} = c$. Moreover, for any $\boldsymbol{\Delta}$ such that $\sum_{i}\Delta_i = 1$, there also exist $H$ such that $\boldsymbol{\Delta} = H^{-1}_{:, 1}$.
\end{customlemma}
\begin{proof}
We aim to choose (column vectors) $\mathbf{u}_1,\cdots, \mathbf{u}_N$ and $\mathbf{v}_1,\cdots, \mathbf{v}_N$ such that $H = \begin{pmatrix}\mathbf{u}_1^\top\\\vdots\\\mathbf{u}_N\end{pmatrix}$ and $H^{-1} = \begin{pmatrix}\mathbf{v}_1,\cdots, \mathbf{v}_N\end{pmatrix}$ satisfy the conditions in the claim. Let $\langle \cdot, \cdot \rangle$ denote the standard dot product in $\R^N$. First choose $\mathbf{u}_1=(1,\cdots, 1)$.

Choose $\mathbf{v}_1^0 \in \R^N$ such that $\langle \mathbf{v}_1^0, \mathbf{u}_1 \rangle = 1$. Let $d = \langle \mathbf{v}_1^0, \boldsymbol{\alpha} \rangle$. If $d=c$, then set $\mathbf{v}_1 = \mathbf{v}_1^0$. Else, take $\mathbf{v} \in \mathbf{u}_1^{\perp}$ and $\mathbf{v} \not \in \boldsymbol{\alpha}^{\perp}$. $\mathbf{v}$ exists since $\boldsymbol{\alpha} \not \propto \mathbf{u}_1$. Choose $\mathbf{v}_1 = \mathbf{v}_1^0 + \frac{c-d}{\langle\mathbf{v}, \boldsymbol{\alpha}\rangle}\cdot\mathbf{v}$, then $\langle \mathbf{v}_1, \boldsymbol{\alpha}\rangle  = \langle \mathbf{v}_1^0, \boldsymbol{\alpha}\rangle + \frac{c-d}{\langle\mathbf{v}, \boldsymbol{\alpha}\rangle}\cdot\langle \mathbf{v},\boldsymbol{\alpha}\rangle = d + c - d = c$.

Now choose $\mathbf{u}_2,\cdots, \mathbf{u}_N$ s.t. i) $\langle \mathbf{u}_i, \mathbf{v}_1\rangle = 0,\; \forall i=2,\cdots, N$, and ii) $\mathbf{u}_2\cdots, \mathbf{u}_N$ are linearly independent. This can be done since ${\rm dim}(\mathbf{v}_1^{\perp}) = N-1$. 

Note also that $\mathbf{u}_1,\cdots, \mathbf{u}_N$ are linearly independent: suppose $\sum_i m_i\mathbf{u}_i = \mathbf{0}$, then $\sum m_i\langle \mathbf{u}_i, \mathbf{v}_1\rangle = 0$. But $\langle \mathbf{u}_i, \mathbf{v}_1 \rangle = 0$ for $i=2,\cdots, N$; therefore, $m_1\langle \mathbf{u}_1, \mathbf{v}_1 \rangle = 0$. Since $\langle \mathbf{u}_1, \mathbf{v}_1 \rangle = 1$, we deduce $m_1 = 0$. But this means $\sum_{i=2}^N m_i \mathbf{u}_i = \mathbf{0}$. Since we know $\mathbf{u}_2,\cdots, \mathbf{u}_N$ are linearly independent, $m_i=0$ for $i=2,\cdots, N$.

For $j=2,\cdots, N$, choose $0\neq \tilde{\mathbf{v}}_j \in \text{Span}\left(\left\{\mathbf{u}_i, i\neq j\right\}\right)^{\perp}$. Since $\left\{\mathbf{u}_i, i\neq j\right\}$ are linearly independent, dim$(\text{Span}\left(\left\{\mathbf{u}_i, i\neq j\right\}\right)^{\perp}) = 1$, and thus $\text{Span}(\tilde{\mathbf{v}}_j) = \text{Span}\left(\left\{\mathbf{u}_i, i\neq j\right\}\right)^{\perp}$. Then $\langle \tilde{\mathbf{v}}_j, \mathbf{u}_i \rangle = 0$ when $i\neq j$. When $i=j$, $
\langle \tilde{\mathbf{v}}_j, \mathbf{u}_j\rangle \neq 0$ since otherwise $\mathbf{u}_j \in \tilde{\mathbf{v}}_j^{\perp} = \text{Span}\left(\left\{\mathbf{u}_i, i\neq j\right\}\right)$. Therefore, choose $\mathbf{v}_j = \frac{\tilde{\mathbf{v}}_j}{\langle \tilde{\mathbf{v}}_j, \mathbf{u}_j\rangle}$. 

The above shows the first statement. 

The second statement is obvious by observing that $\mathbf{v}_1$ can be chosen as $\boldsymbol{\Delta}$ by construction.
\end{proof}

\begin{customlemma}{5}\label{lemma:3}
Let $\mathbf{v} \in \R^{M \times N}$ be a vector with $\sum_{i,j} v_{ij} = 0$, and $\mathbf{u}_i \in \R^{N}$ a vector such that $\sum_{j}u_{i,j} = 0$, where $u_{i,j}$ is the $j$th element of $\mathbf{u}_i$. Suppose $\mathbf{u}_i$ and $\mathbf{v}_i$ are linearly independent, $M\geq 3$, $N \geq 2$, $K \geq 1$. Then there exists a distribution of discrete variables $P(\bar{Y}|X, Z)$ where $|\cX|=M , \; |\cZ|=N , \; |\bar{\cY}|=K $ such that
\begin{enumerate}
    \item $\sum_{i,j=1}^{M,N}P(\bar{Y}|X=x_i,Z=z_j) v_{ij} = \mathbf{0}$.
    \item $\sum_{j}^N P(\bar{Y}|X=x_i, Z=z_j)u_{i,j} \neq \mathbf{0}$.
\end{enumerate}

\end{customlemma}

\begin{proof}

Let $P_{ij}$, $i=1, \cdots, M, \; j=1,\cdots, N$ denote the vector of $P(\bar{Y}|X=x_i,Z=z_j)$.



The case with $v_{ij} = 0$ for all $i,j$ is trivial, so assume that there exist some $(i,j)$ such that $v_{ij} \neq 0$. 

If there is some ${i,j}$ such that $v_{ij} = 0$, then we are done by choosing $P_{i_1,j_1} = P_{i_2,j_2}$ for any $(i_1,j_1), (i_2,j_2) \neq (i,j)$ as any valid probability vector, and choosing $P_{ij}$ to be a different valid probability vector with all elements strictly between $0$ and $1$. For example, we can choose $P_{i_1,j_1} = (1/K, \cdots, 1/K)$, and $P_{ij} = (1/K + \epsilon, 1/K - \epsilon,1/K \cdots, 1/K)$ with a small enough $\epsilon$.

If $v_{ij} \neq 0$ for all $(i,j)$, then clearly $\sum_{(i,j) \in \cI_+} v_{ij} = -\sum_{(i,j) \in \cI_-}v_{ij}$, where $\cI_+$ is the index set for the positive $v_{ij}$'s and $\cI_-$ are the negatives. For ease of notation let $S \defeq \sum_{(i,j) \in \cI_+} v_{ij} = -\sum_{(i,j) \in \cI_-}v_{ij}$.

Now first assign arbitrary probability vectors for each $P_{ij},\; (i,j) \in \cI_+$. Wlog, we can assign different probability vectors to each, ensuring that each element of the vectors are strictly between $0$ and $1$. Then the linear combination $\sum_{(i,j)\in\cI_+, k} v_{ij}P_{k,ij}$ is equal to $S$; here, $P_{k, ij}$ is the $k$th element of the vector of $P_{k,ij}$. Now assign for each $(i,j)\in \cI_-$, $P_{ij} \defeq \frac{1}{S}\sum_{(i,j)\in \cI_+}v_{ij}P_{ij}$. Clearly, every element of $P_{ij}$ is non-negative and all elements add up to 1, so $P_{ij}$ is a probability vector. Moreover, $P(\bar{Y}|X,Z)$ constructed like this will satisfy the requirement that $\sum_{i,j=1}^{M,N}P(\bar{Y}|X=x_i,Z=z_j)v_{ij}=0$. 

Now it remains to check condition 2. Take a fixed $i \in \{1,\cdots, M\}$, if $\sum_{i,j}u_{ij}P_{ij} = \mathbf{0}$, then consider $\mathbf{v}_i$ and $\mathbf{u}_i$. Since they are linearly independent vectors, the linear map induced by the usual inner product with $\mathbf{v}_i$ has a kernel with a subspace of at least dimension one that is not contained in the kernel of the linear map induced by $\mathbf{u}_i$. Let $\mathbf{d}$ be a unit vector in this subspace. Define $P_i = [P_{i1}, \cdots, P_{iN}]$. $P_i$ is a $K\times N$ matrix. Construct $P'_i$ by subtracting a small enough multiple of $\mathbf{d}$ from the first row of $P_i$ and adding the same multiple to the second row of $P_i$. Redefine $P_i := P'_i$, and we will have $\sum_{j}u_{ij}P_{ij} \neq \mathbf{0}$ and $\sum_{i,j}v_{ij}P_{ij} = \mathbf{0}$. Repeat this for every $i=1,\cdots, M$ and conditions 1 and 2 will both be satisfied.

\end{proof}

\section{Examples}\label{sec:examples}
\subsection{Hours of work and pay}
For manual labourers in a manufacturing plant, the number of hours they work ($X_i$) determines their monthly wage ($Y_i$). In particular, the monthly wage ($Y_i$) of a manual labourer is a product of the number of hours they work ($X_i$) and their hourly wage ($\alpha_i$):
\begin{align}
    Y_i &= \alpha_i X_i 
\end{align}
The hourly wage may differ between labourers due to various factors like work experience. To get a high-level overview of the human resources in the manufacturing plant, its business owner keeps track of  total work hours ($\bar{X}$) and total cost of wages ($\bar{Y}$) on a monthly basis. That is, we have a summation coarsening map: 
\begin{align*}
    \bar{X} = \sum X_i, \qquad \bar{Y} = \sum Y_i
\end{align*}
In this scenario, Gaussian distribution reasonably approximates the distribution of the number of hours of work of a manual labourer ($X_i$) as most labourers work around an average number of hours, with few working extra hours more or less. Moreover, it is reasonable to assume that the number of hours of work of each labourer is independent of other labourers. Altogether, we have the linear unconfounded case with independent Gaussian micro-level causes.



\subsection{Genomic micro-array data for prediction of diseases}
Genomic micro-array data contains personal genome information. A micro-array is a rectangular grid where every column contains the genomic expressions of one person (or subject), and every row contains the expressions of one coding gene for every person. Personal genomic expressions as collected in the micro-array are used to predict certain disease types ($Y$). The number of genes ($X_i$) present on a micro-array is vast; moreover, a population of genes could be jointly causing a disease. This may necessitate coarsening in drug design. 

While two drugs may both claim to intervene on the causes of disease $Y$ on a population level, $\pi((X_i)_i)$, it is possible that they impact the individual amounts of each gene expression differently. Moreover, due to the complexity of gene regulatory networks, there could be confounding present at the micro level. Suppose the practitioners are willing to assume a Linear Gaussian model, then our example in \ref{subsec:4.2} suggests that there is a way for the drug to intervene on the macro variable such that the causal influence of the drug can be directly read off from observational data.

\subsection{Political campaigning for votes}
During the running up to the general election, politicians go around the country to campaign for votes. The amount of time they spend campaigning at county $i$ can be written $X_i$. Some (simplistic) political model may assume a linear relationship between the amount of time campaigning in county $i$ and the votes harvested at that county $Y_i$:
\begin{align}
    Y_i &= \alpha_i X_i
\end{align}
At the end of the campaign, the team may calculate how much effort they put in, in terms of time, and how many votes they won in total. 
\begin{align}
    \bar{X} &= \sum X_i, \;\;\; \bar{Y} = \sum Y_i
\end{align}
It will be noticed that different counties responds differently to the campaign, so $\alpha_i$ is not constant. Now, different campaign teams may all decide to increase their campaign time in order to get more votes, but the resultant vote changes may still be different due to the different allocations to each county. 

\section{More Detailed Discussion of Related Work}\label{app:related_work}
\citep{Rubenstein2017} propose exact transformations, following which \citep{beckers19} propose a series of stronger notions of causal abstraction. As illustrated by the example in Example~\ref{ex:ruben}, the condition required to call a macro system an exact transformation of the the fine-grained one is not satisfied even in extremely simple cases. The moral reason being that when micro causal variables are aggregated, the causal relationships between the micro causal variables do not in general align with the aggregation, such that if multiple micro-cause-states get coarsened to the same macro-state, then the corresponding micro-effect-states get coarsened also to the same macro-state.
By contrast, this work explores what happens when the macro system cannot be called an exact-transformation of the micro system.

\citep{Beckers2019ApproximateCA} explores how to quantify the approximation error when the abstraction is not exact. We take a different angle in this work and do not deal with how good our macro-model is at approximating the micro-model. Rather, our work observes that the aggregated system loses resolution on the original, micro-system, and analyses properties when different pairings between the macro- and micro-systems are made. 

A different but closely-related line of work is undertaken by \citep{Chalupka2016MultiLevelCS, Rischel2021CompositionalAE}, where the coarsest possible consistent coarsening is sought. The macro-variable states are constructed as equivalence classes (of micro-variable states) in the paper, and every micro-variable state in the same equivalence class leads to the same micro-variable effect - this ensures that interventions on the equivalence classes are well-defined. Our work lies in the realm beyond the coarsest consistent partition, and in these cases, the macro interventions become ill-defined.

\citet{pmlr-v213-beckers23a} provides a framework to include constraints and structural equation model in a unified model. Macro variables in their framework leads to constraints on the system, and they propose to think of interventions on macro variables as implemented by disconnecting some of the other variables in the constraint from their original structural equations. Different ways of implementations can arise from choosing different variables to disconnect. This is similar to our notion of micro-realisations, and the differences are: 1) \citet{pmlr-v213-beckers23a} do not consider soft micro-level interventions as micro-realisations whereas we do, 2) they do not define causal models over only the macro variables whereas we do; in doing so, we explicate consequences of aggregation such as ambiguity of confounding.

\citet{anand23aaai} describes a coarse-graining of only the causal DAG, and generalises do-calculus to deal with clusters of variables. However, they do not aggregate the variables within the cluster. As a result, ambiguities do not arise from different micro-realisations in the corresponding micro-variables of a particular macro-intervention.

\citet{Rischel2021CompositionalAE} proposes a framework of deriving approximate causal abstractions such that the errors are composable. However, their perspective differs from ours as they consider a probability distribution of different micro realisations of a given macro intervention, and they use this to define their approximation error for a given macro model, whereas we define macro models based on micro-realisations.

\citet{ijcai2023p638zennaro} introduces the pseudo inverse of the function mapping from the micro-variable states and macro-variable states. This allows them to generalise measures of approximation error to be along other paths on the commutative diagram. The pseudo inverse will, given a macro-variable states as input, output a vector of equal weights one for each micro state that gives rise to the macro state. In the case of a uniform distribution over the micro states, their pseudo inverse a special case of our natural micro realisation. Then, they learn their macro relation based on the given pseudo inverse, whereas we consider different macro relations which are entailed by different micro implementations; thus our perspective is complementary to theirs.

Finally, our work inspires assosiation with \citep{Blom2019}, which describes a framework for modelling cyclic causal models, where solutions to the model are the equilibrium states for some initial conditions, and interventions amount to changing these initial conditions. In particular, the causal dynamics of the model are fixed by differential equations. Although some might be tempted to view the relationship between micro and macro variables in our case as cyclic, it is, however, fundamentally different. In our case by specifying the $P^{do}_{X|\;\bar{x}}$ in the macro-intervention, the dynamics between the macro and micro variables are freely customisable by the practitioner, whereas in their case, the dynamics between the macro and micro variables are fixed a priori by some constraints, for example, by the laws of physics. That being said, there could be situations where the dynamics fixed by the a priori constraints, in some canonical way, already gives rise to the confounding-inhibiting interventions that we described.

\end{document}